\documentclass[a4paper]{article}
\usepackage{amsmath,amssymb,amsfonts}
\usepackage{algorithmic}
\usepackage{graphicx}
\usepackage{textcomp}
\usepackage{xcolor}

\usepackage{url}

\usepackage{caption} 
\usepackage{algorithm}
\usepackage{algorithmic}

\usepackage{newfloat}
\usepackage{listings}
\DeclareCaptionStyle{ruled}{labelfont=normalfont,labelsep=colon,strut=off} 
\floatstyle{ruled}
\newfloat{listing}{tb}{lst}{}
\floatname{listing}{Listing}
\usepackage[utf8]{inputenc}
\usepackage{subfigure}
\usepackage{subcaption}
\usepackage{dsfont}
\usepackage{multirow}
\usepackage{amsfonts}
\usepackage{amsmath}
\usepackage{amssymb}
\usepackage{mathtools}
\usepackage{tabularx}
\usepackage{hyperref}
\usepackage{cleveref}

\usepackage{amsthm}
\newtheorem*{proposition*}{Theorem}

\theoremstyle{plain}
\newtheorem{theorem}{Theorem}[section]
\newtheorem{proposition}[theorem]{Theorem}

\theoremstyle{definition}

\theoremstyle{remark}
\newtheorem{remark}[theorem]{Remark}

\begin{document}




\title{Learning with Noisy Labels through Learnable Weighting and Centroid Similarity}



\author{
\textbf{Farooq Ahmad Wani}$^1$~~~   
\textbf{Maria Sofia Bucarelli}$^1$~~~\\
\textbf{Fabrizio Silvestri}$^1$~~~
\smallskip 
\\
$^1$Department of Computer, Control and Management Engineering   \\ Antonio Ruberti,  Sapienza University of Rome, Italy
}

\maketitle

\begin{abstract}
We introduce a novel method for training machine learning models in the presence of noisy labels, which are prevalent in domains such as medical diagnosis and autonomous driving and have the potential to degrade a model's generalization performance. Inspired by established literature that highlights how deep learning models are prone to overfitting to noisy samples in the later epochs of training, we propose a strategic approach. This strategy leverages the distance to class centroids in the latent space and incorporates a discounting mechanism, aiming to diminish the influence of samples that lie distant from all class centroids. By doing so, we effectively counteract the adverse effects of noisy labels. The foundational premise of our approach is the assumption that samples situated further from their respective class centroid in the initial stages of training are more likely to be associated with noise. Our methodology is grounded in robust theoretical principles and has been validated empirically through extensive experiments on several benchmark datasets. Our results show that our method consistently outperforms the existing state-of-the-art techniques, achieving significant improvements in classification accuracy in the presence of noisy labels. The code for our proposed loss function and supplementary materials is available at \url{https://github.com/wanifarooq/NCOD}
\end{abstract}

\\

\section{Introduction}
\label{introduction}

Deep learning models have achieved state-of-the-art performance on various tasks, including image classification \cite{Krizhevsky2012}, natural language processing \cite{Hochreiter1997}, and speech recognition \cite{Deng2013}. However, their performance depends on the quality of the training data, prior studies have demonstrated deep learning model's susceptibility to overfit noisy labels when training data contains them, resulting in a degraded model performance \cite{Zhang2016}.
 
Label noise in training data can occur for various reasons: human error, the ambiguity of the task, or simply due to the existence of multiple correct labels for a single instance. 

In this paper, we address the issue of label noise in training data by proposing a novel training algorithm called NCOD: Noisy Centroids Outlier Discounting. 

NCOD uses two mechanisms. The first, “\textit{class embeddings}” (Section~\ref{sec:class_embedding}), is rooted in the idea that similar samples have identical labels. We represent each class with a latent space vector, comparing each sample's latent representation to its class representation, and use the result as a soft label. As highlighted in \cite{li2020gradient}, early stopping contrasts the influence of noisy labels (see \cref{fig:train_plots_cifar}). To further prevent noise-induced overfitting, we've introduced a learnable parameter, $u$, in our “\textit{outliers discounting}” mechanism (Section \ref{sec:outlier_discounting}). The efficacy of $u$ is both theoretically demonstrated in Theorem \ref{increasing_of_u} and empirically validated.

\textbf{Motivation and Contributions}.
We propose a technique optimized for training machine learning models in the presence of noisy labels. Our method incorporates an innovative outlier discounting parameter, a first in current literature. This parameter's role in deep neural network training is rigorously assessed both theoretically and practically.

Key contributions include:
\begin{itemize}
\item[(i)] A new loss function using the learned similarity between a sample and its class representation as a soft label.

\item[(ii)] A novel, theoretically sound, regularization term used to discern a discount factor for sample noise.

\item[(iii)] Comprehensive testing on datasets with synthetic or natural noise, highlighting our method's edge over SOTA.

\item[(iv)] Our approach eliminates the need to know the dataset's noise rate or use anchor points, facilitating straightforward real-world applications.
\end{itemize}

\section{Related Work}

There is a large body of work devoted to the challenge of learning with noisy labels. Several methods are based on \textbf{robust loss functions}. Symmetric losses have been shown to be noise-tolerant under symmetric or uniform label noise \cite{ghosh2017robust, ghosh2015making}, thus several works introduce symmetrized versions of CE and generalized loss functions \cite{wang2019symmetric, zhang2018generalized, feng2021can}.

Diverse approaches estimate noise transition matrices, which represent probabilities of label changes, or subsequently introduce a mechanism for \textbf{loss correction} \cite{patrini2017making, menon2015learning, zhu2022detecting, hendrycks2018using, zhu2021clusterability}. These methods often rely on assumptions like the diagonal dominance of the matrix, which makes them unsuitable for high-noise settings. Additionally, the often assumed requirement for clean samples further compounds challenges in practical, real-world scenarios.

\textbf{Neighboring-based noise identification} approaches have been explored by several methods \cite{zhu2021clusterability, zhu2022detecting, iscen2022learning, wu2020topological, guo2018curriculumnet, bahri2020deep}. \cite{zhu2021clusterability} and \cite{zhu2022detecting} in particular propose methods based on the cluster ability condition. 
Our method differs from cluster ability techniques by avoiding using neighboring points' labels, focusing instead on their similarity to the mean of points belonging to a specific class.
\cite{iscen2022learning} employs an additional loss term inspired by label propagation in graph-based semi-supervised learning to encourage similar predictions among samples and their neighbors. 

Several approaches exploit in various ways the \textbf{early-learning phenomenon} \cite{arpit2017closer}. 
Some techniques have been proposed to improve the quality of training data by treating small-loss samples as correctly labeled during the training process \cite{gui2021towards,han2018co,pmlr-v119-yao20b,wang2018iterative, xia2021sample}. 
Some methods focus on sample relabelling \cite{reed2014training, arazo2019unsupervised}, for instance, \cite{reed2014training}  introduced the bootstrapping loss, where a fixed linear combination of the annotated label and current prediction serves as the regression target. This method also leverages the fact that correct labels tend to be approximated before incorrect ones. 

Other techniques address noisy label challenges using two networks, splitting the training set and training two models for mutual assessment \cite{jiang2018mentornet, han2018co, li2020dividemix, kim2023crosssplit}. 
For instance, \cite{gui2021towards} trains a classifier on noisy data, identifying cleaner samples through lower losses. They then train a second classifier solely on these small-loss samples. However, challenges remain, such as determining the optimal clean dataset size.
In \cite{jiang2018mentornet}, an additional network is pre-trained to select clean instances, guiding the main network's training. \cite{han2018co} maintains twin networks with matching architectures, each updated using the other's low-loss samples. Sample selection is a critical issue in these methods; discarding too many samples can lead to lower accuracy.  While leveraging the initial phase's clean data utilization, our approach exploits the early learning with a single model, simplifying the training process and reducing the computational resources needed.

Regarding \textbf{regularization} for noisy labels, mixup augmentation \cite{mixup} is a widely used method that generates extra instances through linear interpolation between pairs of samples in both image and label spaces.
\textbf{Label correction} can be considered \cite{liu2020early, chen2023imprecise}.
\cite{liu2020early} study a new strategy for detecting the start of the memorization phase for each category during training and introduce a novel regularization to correct the gradient of the cross-entropy cost function.
\textbf{Reweighting techniques} aim to improve the quality of training data by using adaptive weights to reweight the loss for each sample \cite{liu2015classification, zhang2020decoupling, han2018co,pleiss2020identifying}. For instance, \cite{ren2018learning} proposes to learn adaptive weights based on the gradient directions of the training samples. However, this method needs a clean validation set. 

Our model is related to reweighting techniques but it differs from previous methods in both the learning process of the reweight parameter and its integration into the cross-entropy loss.
Instead of directly reweighting the loss, our method employs the learnable parameter $u$ (explained in Section \ref{sec:outlier_discounting}) that modifies the cross-entropy loss by shifting the prediction of samples. This reweighting mechanism effectively mitigates the influence of noisy samples; the larger values of the learnable parameter the less the network's prediction must align with the class given by the dataset that can be potentially noisy. 

\cite{liu2022robust} also make use of a learnable parameter for the sample starting from the idea that noise is sparse and incoherent with the network learned from clean data, model the noisy labels via another sparse over-parametrized term, and exploit implicit regularization to recover and separate underlying corruptions.

We direct the interested reader to the survey papers of \cite{han2020survey} and \cite{song2022survey} for a comprehensive analysis of the existing literature.
\begin{figure*}
\centering
\resizebox{0.8\textwidth}{!}{
\subfigure[Epoch 1. CE Loss]{
\label{fig:Training_bce_1}
\includegraphics[width=0.45\linewidth]{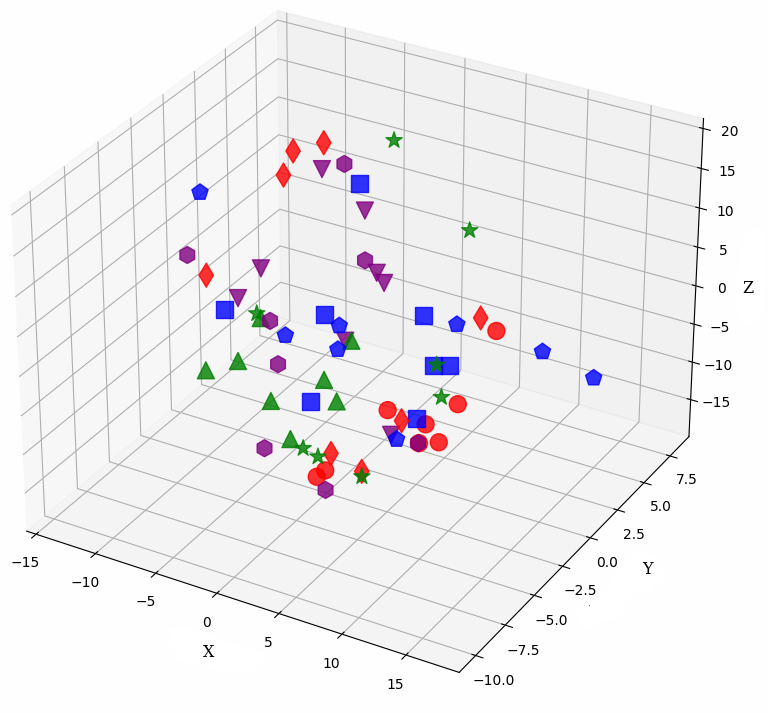}
}\quad
\subfigure[Epoch 14. CE Loss]{
\label{fig:Training_bce_14}
\includegraphics[width=0.45\linewidth]{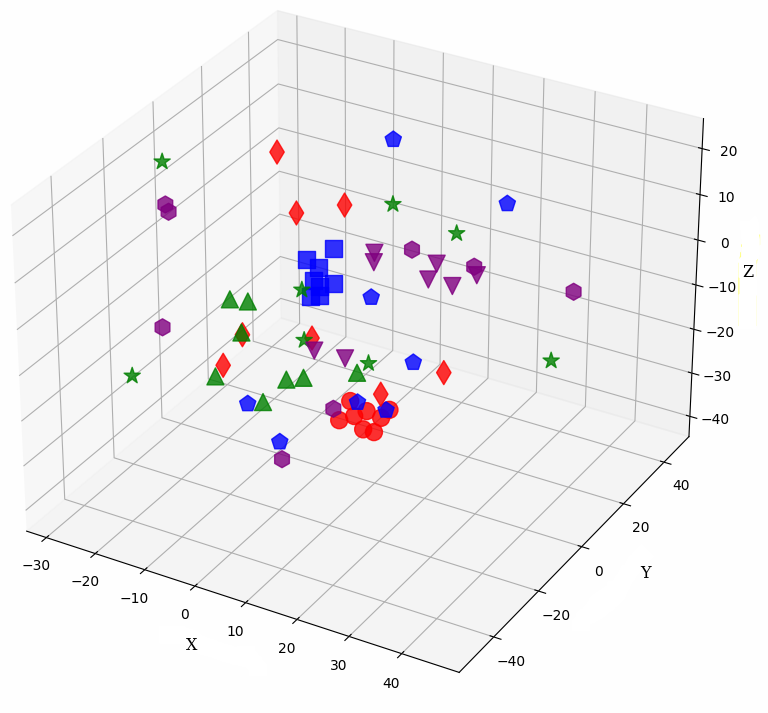}
}\quad
}
\resizebox{0.8\textwidth}{!}{
\subfigure[Epoch 96. CE Loss]{
\label{fig:Training_bce_96}
\includegraphics[width=0.45\linewidth]{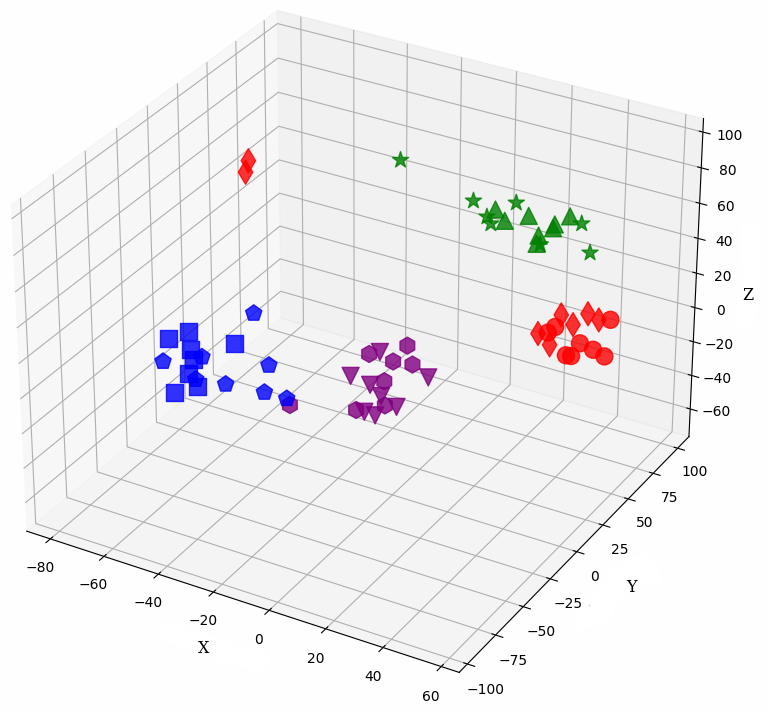}
}\quad
\subfigure[Epoch 96. NCOD Loss]{
\label{fig:Training_ours_96}
\includegraphics[width=0.45\linewidth]{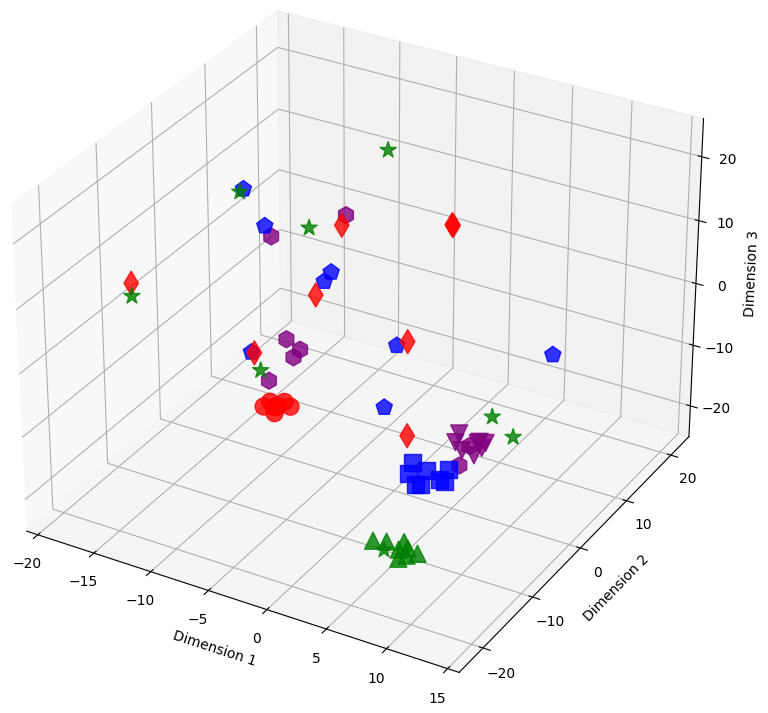}
}\quad
}
 \caption{  Sample embeddings of four classes from CIFAR-100 with 20\% symmetrical noise.
 Colors represent classes, and shapes distinguish noisy and pure samples: Blue (square: pure, pentagon: noisy), Red (circle: pure, diamond: noisy), Green (triangle-up: pure, star: noisy), and Purple (triangle-down: pure, hexagon: noisy).}
 \label{fig:Training_bce_and_ours_20}
\end{figure*}

\section{Evolution of Training}
\label{sec:evolution}
We analyze the training evolution when using our loss function compared to cross entropy loss function.

\subsection{Latent Space Representation}

We examine the evolution of latent representations in the model's penultimate layer throughout the training process. The 3D visualizations in \cref{fig:Training_bce_and_ours_20} illustrate the shifts in these representations for selected training samples from the CIFAR-100 dataset with 20\% symmetrical noise at epochs 1, 14, and 96. These samples come from four different classes. The model uses cross-entropy loss in subfigures (a-c) and our loss, NCOD, in subfigure (d).

In the visualizations, each color corresponds to a distinct class, while different shapes differentiate between noisy and pure samples of the same class. As noted in prior research \cite{liu2020early, gui2021towards, arpit2017closer, wang2018iterative}, the model initially learns from the pure samples of each class. This results in tight clusters of latent representations for those samples. In contrast, noisy samples have latent representations that stand apart from the class they are erroneously associated with in \cref{fig:Training_bce_and_ours_20}. As training advances,  the model also begins to learn from the noisy samples, causing their latent representations to move closer to the center of a cluster of their incorrectly labeled classes. 
To mitigate the impact of label noise, we propose a novel loss function that incorporates an additional parameter $u$ acting as a regularization term. By leveraging this parameter in the loss function, we successfully hinder the model from learning from noisy samples \cref{fig:Training_bce_and_ours_20}. Consequently, the clusters formed by pure samples become more tightly packed, whereas noisy samples remain scattered and distanced when our loss function is employed, as shown in  \cref{fig:Training_ours_96}.
The behavior at epochs 1 and 14 for our loss is similar to that of the CE loss. The relative plots and additional plots on the latent representations can be found in Appendix \ref{appendix:Latent_Representations}. 
 We utilize t-SNE to reduce the high-dimensional embeddings. Despite the reduction in dimensionality, the difference in behavior between the latent representations of noisy and pure samples is still visible.

\subsection{Distribution around Class Cluster Centers}
To enhance our understanding of the observed behavior, depicted in \cref{fig:Training_bce_and_ours_20}, illustrating representation evolution in latent space, we analyzed sample distribution around cluster centers. 

\begin{figure}
\centering
\subfigure[ep. 1. CE Loss]{\label{fig:Training_bce_and_ours_20_d_epoch_1}
\includegraphics[width=0.40\linewidth]{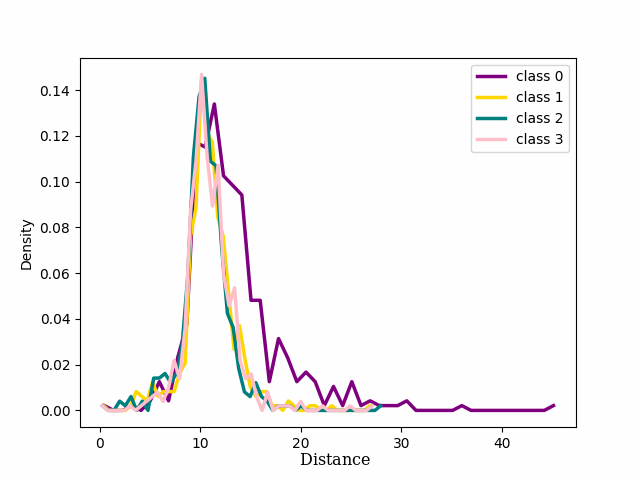}
}\quad
\subfigure[ep. 18. CE Loss]{\label{fig:Training_bce_and_ours_20_d_epoch_18}
\includegraphics[width=0.40\linewidth]{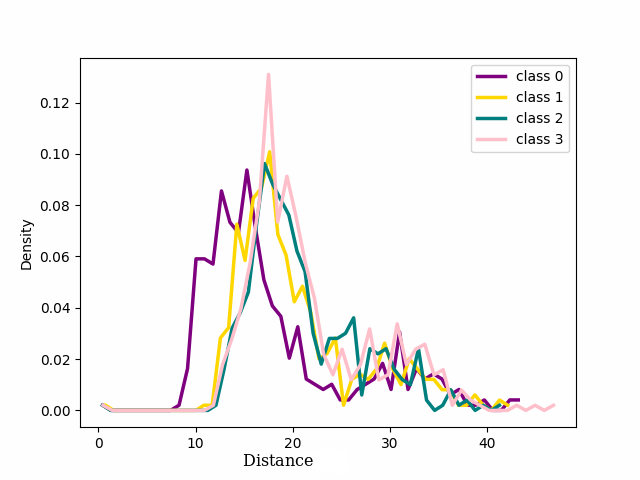}
}\quad
\subfigure[ep. 96. CE Loss]{\label{fig:Training_bce_and_ours_20_d_epoch_96}
\includegraphics[width=0.40\linewidth]{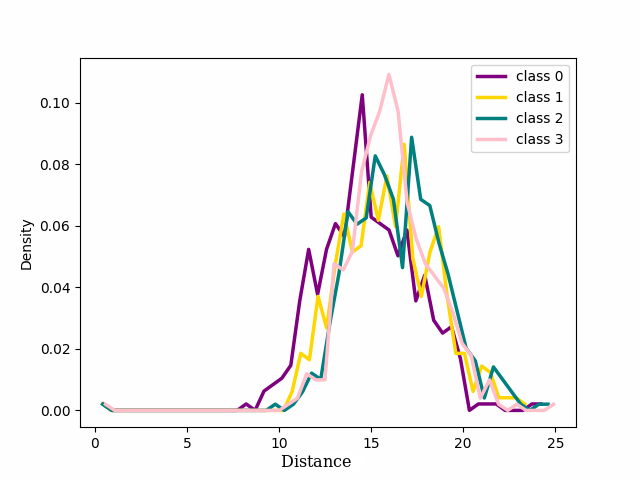}
}\quad
\subfigure[ep. 96. NCOD Loss]{
\label{fig:Training_our_20_d_epoch_96}
\includegraphics[width=0.40\linewidth]{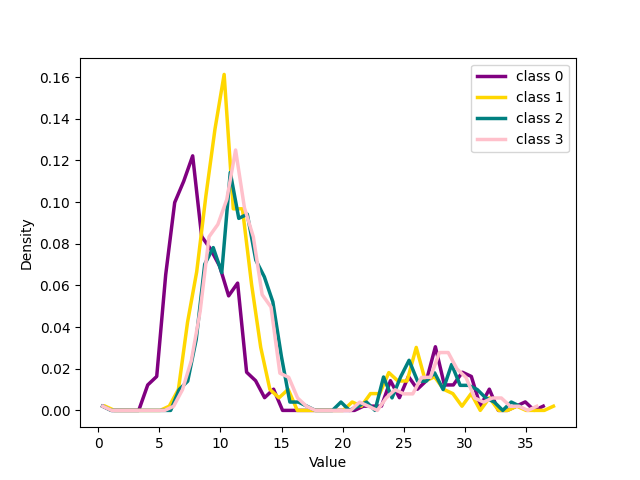}
}\quad
\caption{ Distribution of four classes from CIFAR-100 with 20\% symmetrical noise. ep. is an abbreviation for ``epoch''. 
}
\label{fig:Training_bce_and_ours_20_d}
\end{figure}

\begin{figure}
    \centering
    \subfigure[ep 18. Noisy Samples]{
\label{fig:Distributions_noisy_18}
\includegraphics[width=0.40\linewidth]{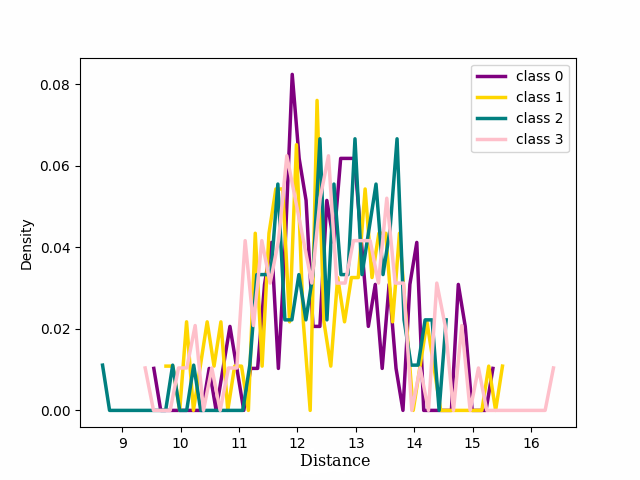}
}\quad
\subfigure[ep. 18. Pure Samples]{
\label{fig:Distributions_noisy_16}
\includegraphics[width=0.40\linewidth]{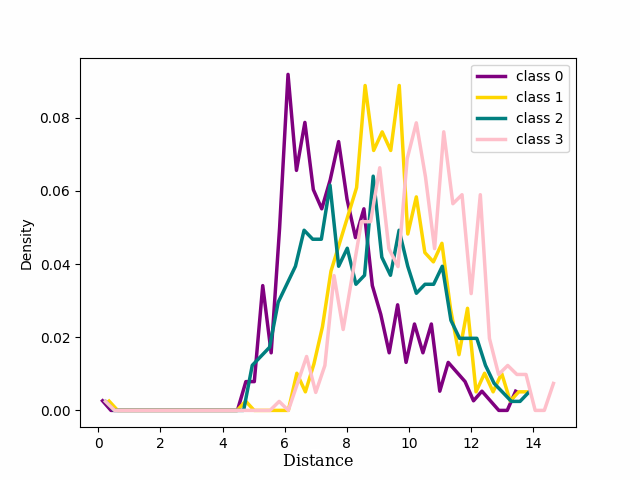}
}\quad
    \subfigure[ep. 96. Noisy Samples]{
\label{fig:Distributions_noisy_96}
\includegraphics[width=0.40\linewidth]{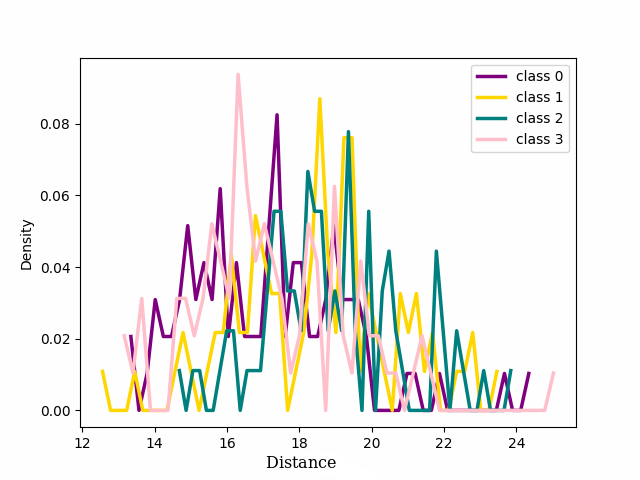}
}\quad
\subfigure[ep. 96. Pure Samples]{
\label{fig:Distributions_pure_96}
\includegraphics[width=0.40\linewidth]{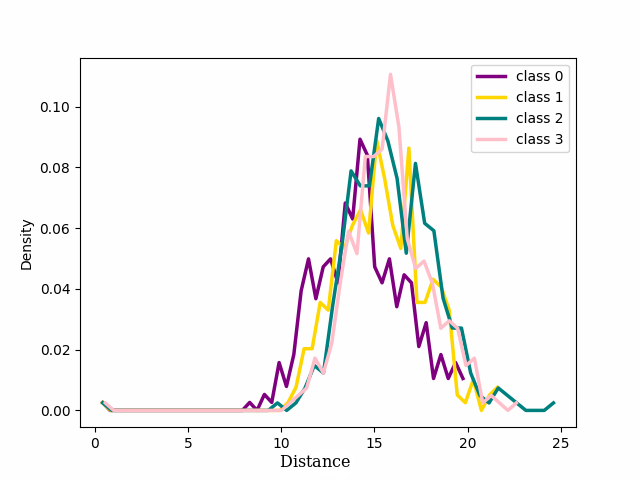}
}
\caption{Distribution of four classes for noisy and clean labels from CIFAR-100 with 20\% symmetrical noise using  CE Loss. ep. is an abbreviation for ``epoch''. 
\label{fig:distributions_noisy_and_clean_ce}
}
\end{figure}

\begin{figure}
    \centering
    \subfigure[ep. 96. Noisy Samples]{
\label{fig:Distributions_noisy_96_ours}
\includegraphics[width=0.40\linewidth]{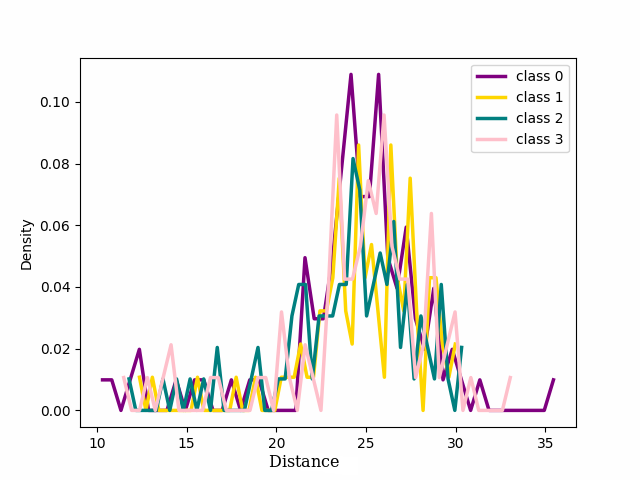}
}\quad
\subfigure[ep. 96. Pure Samples]{
\label{fig:Distributions_pure_96_ours}
\includegraphics[width=0.40\linewidth]{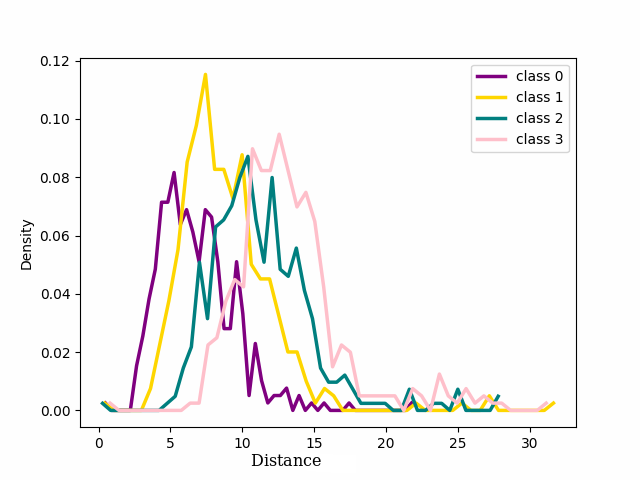}
}\quad
\caption{Distribution of four classes of noisy and clean labels from CIFAR-100 with 20\% symmetrical noise using NCOD Loss. ep. is an abbreviation for ``epoch''.
\label{fig:distributions_noisy_and_clean_ours}
}
\end{figure}

We focused on the distribution of samples from each class around their cluster centers, which we refer to as the `seed' for the rest of our discussion. The seeds are selected as the point in a class that has the minimum Euclidean distance to every other point within the same class. \cref{fig:Training_bce_and_ours_20_d} shows the distribution of the distance in the latent space between the seed for each of the four classes chosen for \cref{fig:Training_bce_and_ours_20} and the other samples that in the training dataset belong to those classes. 
At the beginning of training, cross entropy loss results in a unimodal distribution, (\cref{fig:Training_bce_and_ours_20_d_epoch_1})  shifting to bimodal (\cref{fig:Training_bce_and_ours_20_d_epoch_18}) as the model learns from pure samples.
It means that we can now distinguish between two groups of samples that behave differently.
These two groups reveal to be exactly the pure and noisy samples as can be seen in \cref{fig:distributions_noisy_and_clean_ce,fig:distributions_noisy_and_clean_ours}. As evident from the figure the distribution of noisy and clean samples aligns with the distinct modes observed in cases where the distribution is bimodal.
However, overfitting on noisy samples reverted the distribution to unimodal (\cref{fig:Training_bce_and_ours_20_d_epoch_96}). 

Contrarily, with our loss function once the distribution becomes bimodal, the bimodal nature of the distribution persisted (\cref{fig:Training_our_20_d_epoch_96} and Appendix \ref{appendix:distributions}). Enlarged plots have been included in the Appendix \ref{appendix:distributions}.
These results align with the findings presented in \cref{fig:Training_bce_and_ours_20}, providing quantitative insight into the phenomenon.

To demonstrate the consistency of our analysis across different datasets and architectures, we repeated the same analysis for the MNIST dataset using CNN. The results can be found in Appendix \ref{appendix:distributions}.


\section{Preliminaries}

In this section, we formulate the task of learning with noisy labels and define the notation used in this paper. 

Consider a noisy labeled dataset, $D = \{(x_i, y_i) \mid i \in [1, n] \}$, where $x_i$ denotes the $i$-th training sample and $y_i$ its associated label. Here, $C$ represents the number of classes. In our notation, $y_i$ is a one-hot encoded vector of dimension $C$. Additionally, we'll use $c_i$ to represent the class of sample $i$, specifically $c_i \in \{1, \dots, C\}$. In this setting, the corruption status of each training pair $(x_i, y_i)$ and the fraction of mislabeled samples in the training dataset remain unknown.
The goal is to find the best model 
$ f^*(\theta, x_i): R^d \rightarrow R^C $
that learns only from the true labeled data in our training data and ignores the noisy labels. 

 We denote by $\phi(x_i)$ the output of the second to the last layer of our network, i.e., the embedding of $x_i$ in the feature, or latent space, while $h_i = \frac{\phi(x_i)}{||\phi(x_i)||}$\footnote{Hereinafter, the symbol $|| .|| $ denote the $l_2 $ norm, i.e. $ || x|| = \sqrt{\sum_{i=1}^d x_i^2 }$}. With $\hat{c}_i$ we denote the class predicted by the network, i.e. $\hat{c}_i :=\underset{j = 1, \dots, C}{\operatorname{argmax}} f( \theta,x_i)_j$.
Moreover, we will denote by $ \hat{y}_i $, the one-hot vector relative to the class predicted by the network. 

\section{Methodology}
Our proposed method utilizes the intrinsic relationships between representations of data points belonging to the same class and the phenomenon that the model tends to learn from clean samples first.
To address the issue of overfitting due to noisy labels, we introduced a learnable parameter $u \in [0,1]^n$ of size $n$, i.e., the number of samples in the training dataset. Specifically, for each sample $i$, we have a corresponding parameter $u_i$, which serves as an indicator of the level of noise present in the sample $i$. 
Both introduced novelties ensure that the importance of noisy labels gradually decreases as training progresses, while clean data becomes more relevant.
 To learn the network weights $\theta$ and the parameter $u$, we use the following loss functions $\mathcal{L}_1$ and $\mathcal{L}_2$.
 

\begin{equation}
\begin{split}
\label{eq:loss}
&\mathcal{L}_1(\theta, u_i,x_i, y_i) =   \mathcal{L}_{\text{CE}} (f(\theta,x_i) + u_i\textcolor{black}{ \cdot y_i}, \tilde{y}_i) , \\
&\mathcal{L}_2 (\theta, u_i,x_i, y_i) =  \frac{1}{C} || \hat{y
}_i +u_i\textcolor{black}{ \cdot y_i}  -{y}_i ||^2_2
\end{split}
\end{equation}

Note that the predicted class $\hat{y}_i$ is dependent on the parameters of the network, $\theta$. While
$\tilde{y}_i$  denotes the soft label for sample $i$. The soft label is obtained as the similarity of the sample's features with respect to class centroids, see Section  \ref{sec:class_embedding}.
The final losses will be the mean of $\mathcal{L}_j(\theta, u, x_i, y_i)$ over the samples for both $j=1,2$, denoted by $\overline{\mathcal{L}_j}$.
We used stochastic gradient descent to update the parameters,  performing back-propagation through the parameters $\theta$ in the loss $\mathcal{L}_1$, and $u$ in the second loss function $\mathcal{L}_2$.
Specifically, we update the parameter using the following procedure:
\begin{align}\label{eq:joint_update}
\theta^{t+1} &=\theta^{t} - \alpha \nabla_{\theta} \overline{\mathcal{L}_1}(\theta^{t}, u_i^{t})\\
u^{t+1} &= u^t - \beta \nabla_{u} \overline{\mathcal{L}_2}( \theta^{t}, u_i^{t}).
\end{align}




\textcolor{black}{
The $\mathcal{L}_1$ loss is the cross entropy with soft labels and samples are reweighed according to the parameter $u_i$.
Effective $u_i$ weights should give more importance to clean samples.
Now, the challenge lies in learning these $u_i$ weights.
$\mathcal{L}_2$ leverages the fact that at the beginning, the network makes more errors in predictions for noisy samples. 
Theorem  \ref{increasing_of_u} shows that, for each sample, the updated parameter $u_{t+1}$ decreases or remains constant at zero when the network prediction is correct, while it increases when the predictions are incorrect. In our approach, we initialize the parameter $u$ close to $0$. 
Consequently, as shown also in the experiments  (\cref{fig:sample_noise} and \cref{fig:sample_sim} in the Appendix), the $u_i$ parameter grows more rapidly for noisy samples.
This provides a valid rationale for employing the $\mathcal{L}_2$ loss to learn the $u$ parameter. Note that $\mathcal{L}_2$, depending on model predictions, relies on $\theta$. Similarly, $\mathcal{L}_1$ depends on the parameter $u$, which precisely serves as the weight for sample reweighing. However, in $\mathcal{L}_1$, $u$ is used for reweighing purposes and is not actively learned during the process.} The following two sections provide a detailed explanation of the contributions made by the two novel concepts introduced in this work, namely the use of similarity with class centroids as soft labels and the learnable parameter $u$.

\subsection{Class Embeddings}
\label{sec:class_embedding}
We leverage the natural assumption that a sample close to many samples from a given class is likelier to be part of that class than not. Therefore, for each data sample, we consider two labels. The label from the dataset (which can be noisy) and the (soft) label from close data points. 



We propose to use the similarity between a sample and a summary of all the samples in the training dataset belonging to the same class as a soft label.
In particular, for each class $c \in \{1, \dots C\}$, we construct its representation by averaging all the latent representations of that class's training samples. 
In our case, the latent representation of the class $c \in \{1, \ldots, C\}$ is $\phi(c) = \frac{1}{n_c}\sum\limits_{i=1, y_i=c}^n \phi(x_i)$. 
Here, $n$ is the size of the training set, and $n_c$ is the number of samples belonging to class $c$.
We denote by $\bar h_c$ the vector obtained normalizing $\phi(c)$, 
which we refer to as the class embedding:

$$ \bar h_c = \frac{\phi(c)}{||\phi(c)||}    \text { for } c  \in \{ 1, \dots C\}$$

The soft label for the $i$-th sample, denoted by $\tilde{y}_i$, is the similarity between its normalized feature representation $h_i$ and class embedding $\bar h_{c_i}$,  where $c_i$ is the sample's label:
 $$\tilde{y}_i= [\max({h_i \bar h_{c_i}^T},0) y_i] $$ 
The $C$-dimensional vector $\tilde{y}_i$ has a single non-zero entry, aligned with the non-null entry of $y_i$, with a value equal to the maximum between zero and the sample's similarity to its class embedding. This ensures that the soft labels are non-negative values, preserving the convexity of the cross-entropy loss. The soft label assigns more weight to samples closer to their class centroids and less weight to those farther away.

To determine centroids $\phi(c)$, we also leveraged the parameter $u$, where we gradually decreased the number of $c$-labeled samples for $\phi(c)$ calculation as epochs progressed. Samples are selected from the subset with the lowest values of $u$.

As the model learned to differentiate between noisy and clean samples, we computed class representation using samples with the highest clean probability, exploiting the model's discrimination ability for enhanced efficiency and accuracy. We progressively reduce the utilization of samples labeled as $c$ for computing $\phi(c)$ as epochs unfold so that only $50\%$ of these samples are employed in the final epoch. This choice accounts for an unknown noise percentage; with a known percentage, this should align with the fraction of clean data.




\subsection{Outlier Discounting}\label{sec:outlier_discounting}

We introduce the parameter $u$ to leverage the observed phenomenon that deep networks, despite their capacity to memorize noisy data \cite{Zhang2016}, have been shown in several studies to prioritize learning from the clean samples, i.e. simpler patterns, first \cite{arpit2017closer,  wang2018iterative, li2020gradient, liu2020early, gui2021towards, pmlr-v97-shen19e}. Loss $\mathcal{L}_2$, used to learn this parameter, exploits precisely the observation that during the early stages of training the model tends to make more accurate predictions for clean samples and more erroneous predictions for noisy samples.

Theorem  \ref{increasing_of_u} shows that, for a particular sample $i$, the updated  parameter $u_i^{t+1}$ decreases or remains constant at zero when the network prediction is correct while  it increases with respect to the value it had at epoch $t$ when the predictions are inaccurate.
In our approach, we initialize the parameter $u$ using a normal distribution, with very small mean and variance parameters (the actual values are $\mathcal{N}(1 e-8, 1e-9)$). 
Since, at the initial stages of training, the network usually provides more accurate predictions for clean samples and less accurate predictions for noisy samples, the parameter $u_i$ grows faster for noisy samples and at a slower rate for clean samples. This allows our method not to start overfitting the noisy data at later stages of the training, as shown in Section~\ref{sec:evolution}.

\begin{proposition} \label{increasing_of_u}
Let $(x_i, y_i)$ be a sample, $\theta^t$ be the parameters of the network at epoch $t$, and $u^t_i$ be the parameter for outlier discounting relative to sample $i$ at epoch $t$. Let $ \hat{c}^t_i$ be the prediction of the network at time $t$, and $y_i$ be the class of sample $i$.
Suppose $\hat{c}^t_i = y_i$, then $u^{t+1}_i < u^{t}_i$ when $u^{t}_i \neq 0$, and $u^{t+1}_i =0 $ otherwise. 
When $\hat{c}^t_i \neq y_i$, $u^{t+1}_i \geq u^{t}_i$.
Moreover $u_i^t<1$ for $t \in \mathbb{N}$. 
\end{proposition}



\begin{proof}

 The update rule for $u $ will be
\begin{equation}
    \begin{split}
        u^{t+1}_i &= u^{t}_i - \beta \; \partial_{u_i} \mathcal{L}_2  = u^t_i - 2\frac{\beta}{C} ( \delta_{c_i,\hat{c}_i(t)} -1 + u^t_i)
    \end{split}
\end{equation}
$\delta_{c_i,\hat{c}_i}$ equals 1 if the predicted label matches the sample label in the dataset, and 0 otherwise.
So the update becomes:
\begin{equation}
\label{eq:teo}
 u^{t+1}_i  = \begin{cases}
 u^t_i(1 - 2\frac{\beta}{C}) \text{ if } \hat{c}_i(t)= c_i  \\
u^t_i(1 - 2\frac{\beta}{C}) + 2\frac{\beta}{C} \text{ if }  \hat{c}_i(t) \neq c_i 
 \end{cases}
\end{equation}
In our setting $\beta$ is the learning rate for $u$ and $C$ the number of classes, so $2\frac{\beta}{C} <1$.
From  equation \ref{eq:teo} it follows that if $u^0_i = 0$, $$u^{t+1}_i = 2 \frac{\beta}{C}\sum_{k=0}^t \left( 1- 2 \frac{\beta}{C} \right)^{k} \left(1 - \delta_{\hat{c}_i(t-k),c_i} \right) $$
We can see that the maximum of the sum is reached when predictions are always wrong. In this case, the sum becomes a geometric sum with a ratio smaller than one, and we obtain $\lim_{t \to \infty}u_i^{t} = 1$.
If the classes coincide, $ u^{t}$ is multiplied by a value less than one, so it either decreases in magnitude or remains at zero if its initial value was zero.
If the prediction is incorrect, since $u_i^{t}<1$, $u^{t+1}$ increases.  Indeed  $ u^t_i(1 - 2\frac{\beta}{C}) + 2\frac{\beta}{C} > u^t_i$ if and only in $u^t_i<1$.
\end{proof}

\begin{figure}[ht]
    \centering
    \includegraphics[width=\linewidth]{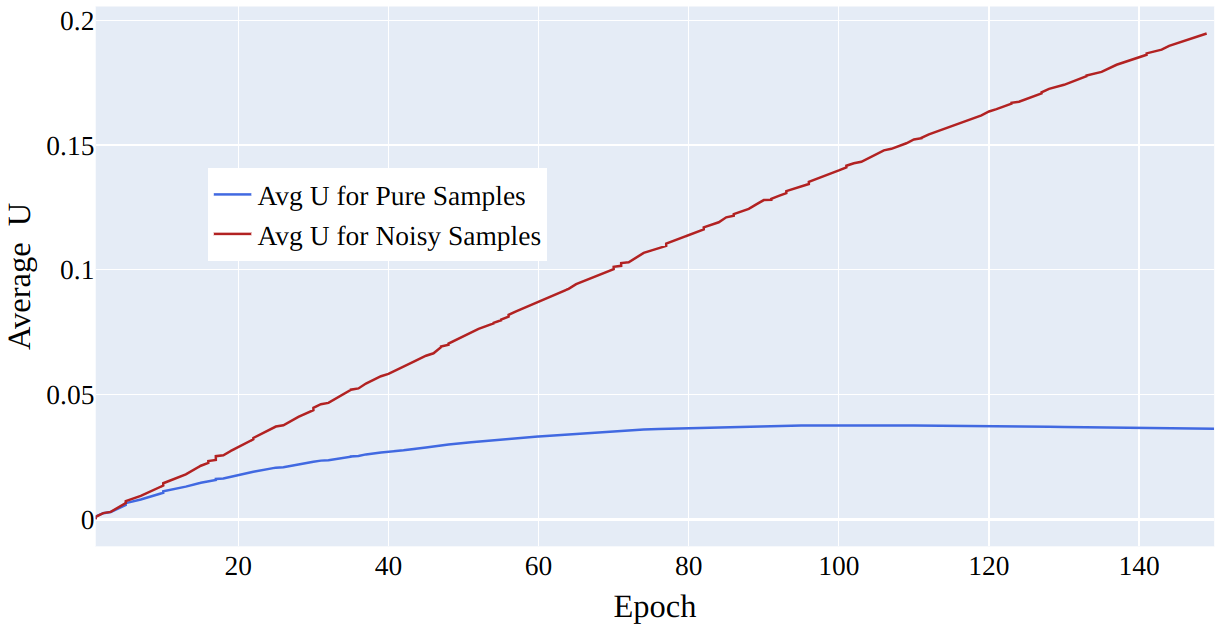}
    \caption{
    CIFAR 100 $50\%$ symmetrical noise.}
    \label{fig:avg_u}
\end{figure}

In \cref{fig:avg_u}, we observe the correlation between the average value for the $u$ parameter and the fact that a sample is noisy. the average of $u$ remains small for clean samples, while it grows more rapidly for noisy samples. 

The parameter $u_i$ affects the loss term $\mathcal{L}_1$ since in the cross entropy loss the prediction is shifted by $u_i$. $u$ serves as a mechanism for mitigating the impact of noisy samples. The loss associated with sample $i$ is $- \log(f(\theta,x_i)+u_i)$, this component will decrease as $u_i$ grows, leading to a decreased impact of the sample on the total loss.
The samples with large $u_i$ are given less weight. We proved that, since the parameter $u$ is smaller than $1$ (see Theorem  \ref{increasing_of_u}), 
it is more effective to reduce the weight of samples with large $u$ by inserting it inside the logarithm rather than multiplying the corresponding loss term by $(1-u)$, see Remark \ref{remark_position_of_u}.  We emphasize that the length of the vector $u$  is equal to the number of samples. Since it is updated only once per epoch, it does not significantly increase the computational cost of the training process.

\begin{remark}\label{remark_position_of_u}
$-\log(x+c)  < -(1-c)\log(x)$ $ \forall c<1$.
\end{remark}

\begin{proof}
    For $0< c <1$, the function $f(x) = \log(x+c) - (1-c)\log(x)$  has a minimum in  $x = 1-c$.
\end{proof}


We denote our training approach, using the loss function outlined in Equation \cref{eq:loss}, as NCOD. Furthermore, adhering to established literature guidelines \cite{liu2022robust,berthelot2019mixmatch,  li2019learning,tanaka2018joint}, we implemented data augmentation alongside two prevalent regularization techniques. These include the Kullback-Leibler divergence between the model's predictions on original and augmented images and a class balance regularization term to prevent biased predictions towards any single class. This refined loss function is denoted as NCOD+.


\subsection{Cross-Dataset Noise Analysis of Loss Functions}
\begin{figure*}
\centering
\includegraphics[width=0.95\linewidth]{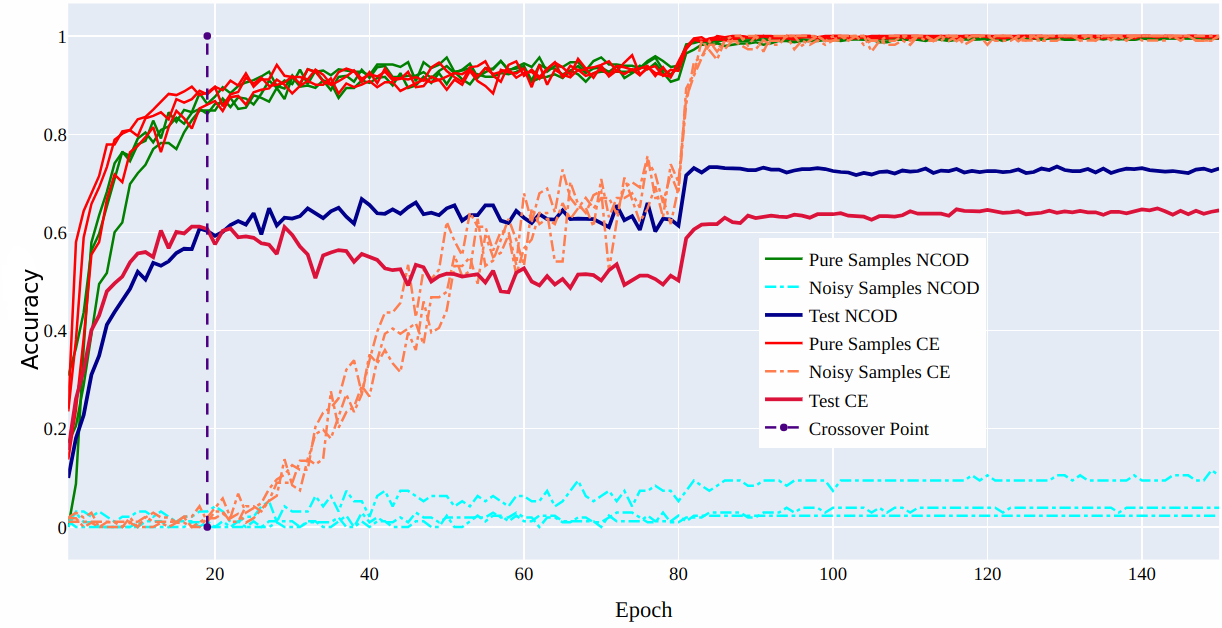}
\caption{  Test accuracy using (NCOD) and Cross entropy (CE) for CIFAR 100 with 20\% symmetric noise.
}
\label{fig:train_plots_cifar}
\end{figure*}

In \cref{fig:train_plots_cifar} we compared the results of our loss function with those of the cross entropy loss on the CIFAR 100 dataset, with 20\% symmetrical noise. Our findings indicate a clear variation in training accuracy between pure and noisy samples. Initially, till epoch 20, pure samples demonstrate higher training accuracy, whereas noisy samples exhibit significantly lower accuracy for both losses. When employing the cross-entropy method, we observed a diminishing gap in training accuracies as the training progressed. This phenomenon contributed to a decline in test accuracy. Conversely, utilizing our loss function, the gap between pure and noisy sample training accuracy remained consistent or widened. This dynamic led to an enhancement in test accuracy. 


\noindent \textsc{\textbf{Ablation study}}. Given the introduction of two novel terms, namely soft labels based on similarity with class centroids and the outlier discounting parameter, we aim to analyze their respective impact on network learning.

\begin{figure}[ht]
    \centering
    \includegraphics[width= \linewidth]{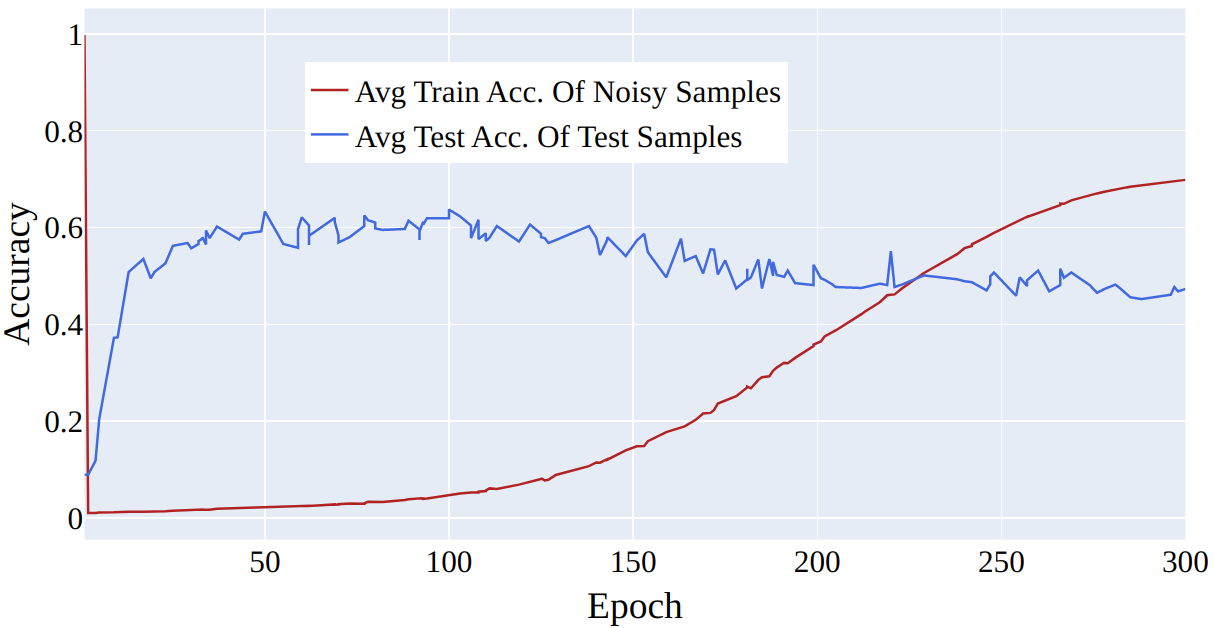}
    \caption{Average test accuracy and average train accuracy on noisy samples for CIFAR $100$ with $50 \% $ of symmetrical noise using only soft labels without not $u$. 
    }
    \label{fig:Noise_test_acc}
\end{figure}

\noindent \textbf{Effect of the Outlier Discounting Parameter on Preventing Overfitting}. We observed that incorporating only soft labels is insufficient for preventing overfitting without using the outlier discounting parameter, as shown in \cref{fig:Noise_test_acc}.
Using the similarity between a sample feature and the class embedding has shown promising results only when the number of training epochs was low. In subsequent epochs, indeed, we observed a decrease in the model's accuracy in the validation set (see \cref{fig:Noise_test_acc}).

\begin{figure}[ht]
    \centering
    \includegraphics[width=\linewidth]{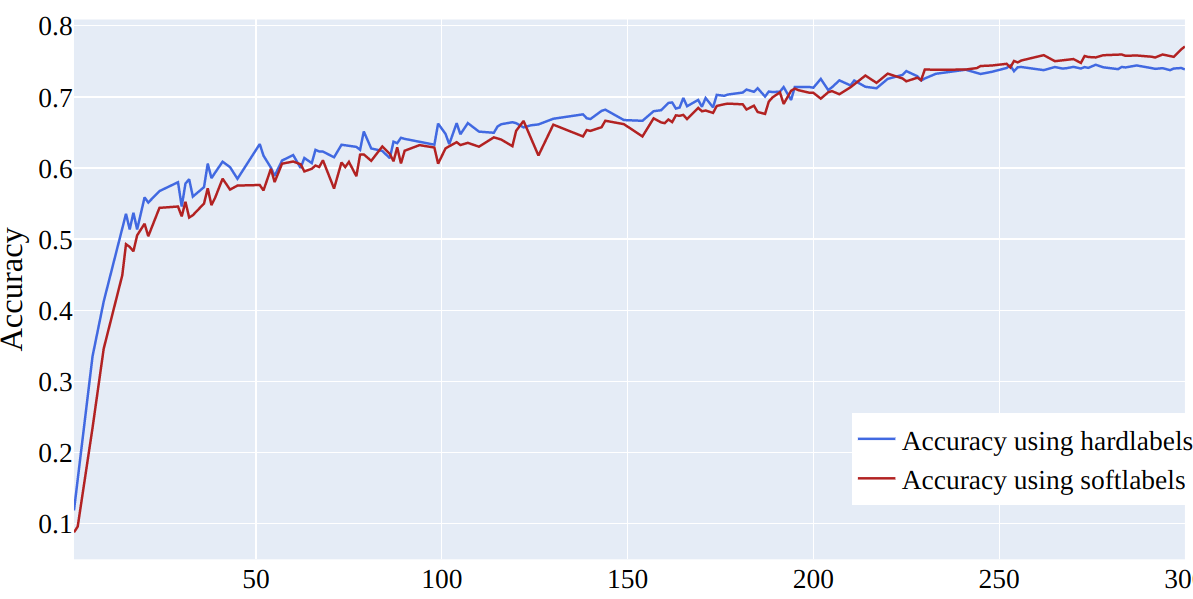}
    \caption{Test accuracy with hard labels or our introduced soft labels  for CIFAR 100 with $50 \%$, symmetric noise. 
    }
    \label{fig:only_u}
\end{figure}

\noindent
\begin{figure}[ht]
    \centering
    \includegraphics[width=\linewidth]{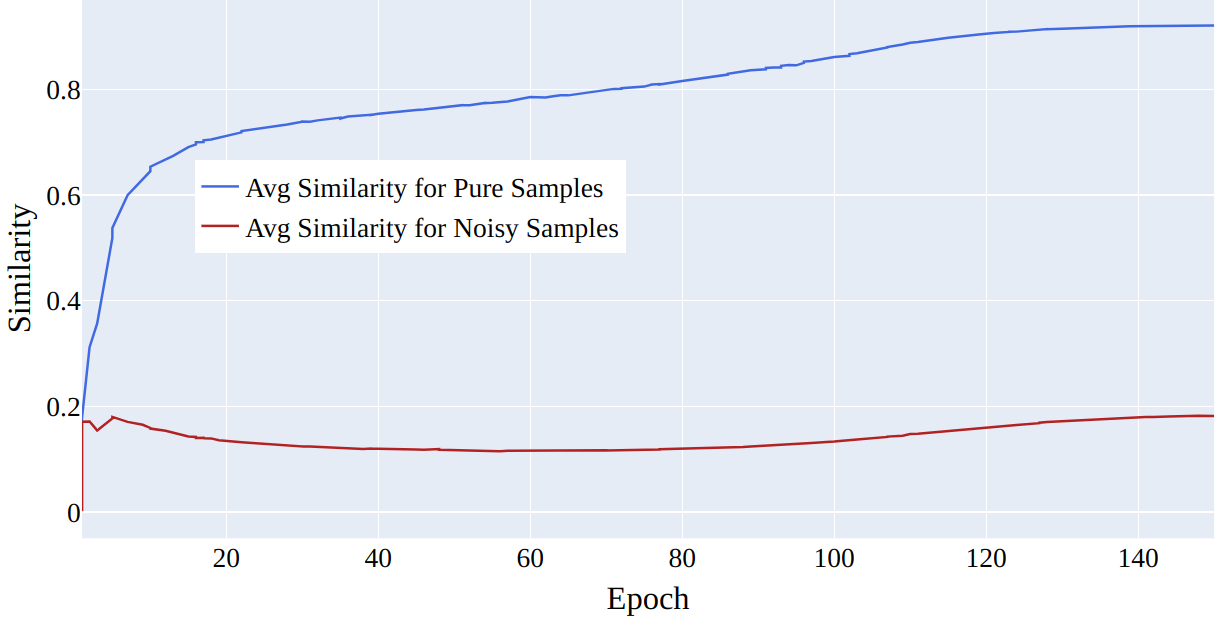}
    \caption{Average similarity for clean and noisy samples. The similarity for pure samples increases as epochs increase while for noisy samples, it remains low without notable improvement. CIFAR-100 $50\%$ symmetrical noise. }
    \label{fig:Figure_u}
\end{figure}
\noindent



\noindent\textbf{Impact of Soft Labels Based on Class Centroid Similarity}.
Our experiments examined the outlier discounting mechanism's effectiveness with hard and soft labels. In \cref{fig:only_u} we show the results of testing the outlier discounting parameter $u$ in combination with soft and hard labels respectively. It is shown that with hard labels the accuracy improves rapidly due to the strong influence of clean samples in early training. Soft labels assign lower weights to distant samples from centroids, reducing their impact. On the other hand, hard labels give uniform weights regardless of distance from centroids. Thus, using soft labels alongside the outlier discounting parameter further diminishes the influence of distant, noisy samples on training. \cref{fig:Figure_u} demonstrates that the final model discerns representations such that noisy samples exhibit lower similarities to their class, whereas clean samples display higher similarities. 


\section{Experiments\label{section:experiments}} 
We conducted experiments under varied settings and compared our method to state-of-the-art techniques with similar configurations. In our experiments, we demonstrate the effectiveness of our method on datasets with both synthetic and realistic label noise. The results of our method are presented in \cref{tab:ensemble100}, \cref{tab:simple100}, and  \cref{tab:miniwebvison}. In this section, we detail the process of performing the experiments on datasets with realistic label noise and datasets with synthetic label noise. 

We compare our findings with results from other methodologies, including those utilizing transition matrix estimation (Forward \cite{patrini2017making}), loss function design (GCE \cite{zhang2018generalized} and SL \cite{wang2019symmetric}), dual-network training (DivideMix \cite{li2020dividemix} and MixUp \cite{mixup}), label refinement (ELR \cite{liu2020early}, \cite{chen2023imprecise}, and SOP \cite{liu2022robust}) and meta-learning based label correction strategy \cite{tu2023LearningFN}.

\noindent \textsc{\textbf{Datasets}}. Our experiments use datasets with both synthetic and realistic label noise. The CIFAR 100 dataset (\cite{articleCifar100}), which has 50,000 training images and 10,000 test images of size 32*32, was used to generate synthetic label noise following the methodologies \cite{https://doi.org/10.48550/arxiv.1804.06872},  \cite{liu2020early}, and \cite{xia2021robust} which are considered to be standard techniques for generating synthetic label noise, it allows us to have a fair comparison of results with other works that have used the same techniques. Symmetric label noise was generated by randomly flipping a certain percentage of labels across all classes, while asymmetric label noise was generated by randomly flipping labels for specific pairs of classes. To compare our results with realistic label noise datasets, we also tested on the mini web-vision dataset \cite{webvision}, which has 65,945 training images with 50 classes and 50 validation images per class. The image sizes vary. All the information about the Network structures and hyperparameters can be found on the GitHub page \url{https://github.com/wanifarooq/NCOD}.

\noindent

\noindent\textsc{\textbf{Experimental results}}. We evaluated our method with ensembled networks \cref{tab:ensemble100},  
single architecture networks \cref{tab:simple100}, and pretrained model architectures \cref{tab:miniwebvison}, which are all commonly used settings. Our method can be effectively applied across a wide range of network configurations.

\begin{table}[ht]
\centering
\begin{tabular}{l|l|l|l|l|l} 
\hline
\multirow{2}{*}{Method} & \multicolumn{3}{l}{Symmetric} & Asym  
& \multirow{2}{*}{Arch}  \\ 
\cline{2-5}
                        & 20\%  & 50\%  & 80\%           & 40\%  
                        &                           \\ 
\hline
CE                      & 58.1  & 47.1  & 23.8           & 43.3  
& R34                  \\ 
MixUp                   & 69.9  & 57.3  & 33.6           & ~-~   
& R34                  \\ 
DivideMix               & 77.1  & 74.6  & 60.2           & 72.1  
& R34                  \\ 
ELR+                    & 77.7  & 73.8  & 60.8           & 77.5  
& R34                  \\

RDA                    & ~-~ & 59.83 & ~-~             & 69.62  
& PR18            \\ 

ILL                    & 77.49 & 75.51 & 66.46          & 75.82  
& PR18            \\ 

SOP+                    & 78.8  & 75.9  & 63.3           & 78.0  
& PR18            \\

SOP+  *                  & 78.23 & 74.83 & 58.83          & 78.1  
& PR18            \\ 

CrossSplit              & 79.9 & 75.7 & 64.6             & 76.8  
& PR18            \\ 

DMLP         & 79.9 & 76.8 & \textbf{68.6}             & ~-~  
& PR18            \\ 

NCOD+                & \textbf{80.34}
& \textbf{77.96}
& 67.9
& \textbf{78.66}
& PR18            \\ 
              & \footnotesize{$\pm$} \footnotesize{0.24} & \footnotesize{$\pm$} \footnotesize{0.56} & \footnotesize{$\pm$} \footnotesize{0.51}         & \footnotesize{$\pm$} \footnotesize{0.234} 
&  \\ 
SOP+ *                   & 78.49 & 76.43 & 61.96          & 78.51      
& PR34            \\ 
CrossSplit              & 81.4 & 77.2  & 67.0             & 79.1  
& PR34            \\ 
NCOD+            & \textbf{81.66}
& \textbf{78.0} 
& \textbf{68.4} 
& \textbf{79.18} 
& PR34            \\
           & \footnotesize{$\pm$} \footnotesize{0.19} &  \footnotesize{$\pm$} \footnotesize{0.58}   & \footnotesize{$\pm$} \footnotesize{0.24}           & \footnotesize{$\pm$} \footnotesize{0.08} 
&         \\
\end{tabular}
\caption{Comparison with the SOTA methods for CIFAR 100 using the two networked ensembles architecture of self-learning under the symmetrical and asymmetrical noise. Mean and standard deviation are recorded over 5 runs. Best performances are highlighted in bold. * denotes experiments conducted with official implementation. The last column shows the architecture, 
R34 stays for Resnet34, PR18 stays for PreActResNet18 while PR34 stays for PreActResNet34. 
}
\label{tab:ensemble100}
\end{table}
\begin{table}[!ht]
\centering
\begin{tabular}{l|l|l|l|l} 
\hline
\multirow{2}{*}{Method} & \multicolumn{4}{l}{Symmetric Noise} 
\\ 
\cline{2-5}
                        & 20\%  & 40\%  & 60\%  & 80\%                                                                                               
                        \\ 
\hline
CE                      & 51.43 & 45.23 & 36.31 & 20.23                                                                       
\\ 
Forward                 & 39.19 & 31.05 & 19.12 & ~8.99~                                                                                   
\\ 
GCE                     & 66.81 & 61.77 & 53.16 & 29.16                                                                                    
\\ 
SL                      & 70.38 & 62.27 & 54.82 & 25.91                                                                                      
\\ 
ELR                     & 74.21 & 68.28 & 59.28 & 29.78                                                                                          
\\ 
SOP 45k~                    & 74.67 & 70.12 & 60.26 & 30.20                                                                                         
\\ 
SOP 45k*                    & 70.4  & 67.7  & 57.1  & 29.26                                                                                       
\\ 
NCOD  45k                   & 73.39
& \textbf{70.14}
& \textbf{61.3}
& \textbf{37.54}
\\ 
              & {$\pm$} \footnotesize{0.15} & {$\pm$} \footnotesize{0.44} & \footnotesize{$\pm$} \footnotesize{0.39}  & \footnotesize{$\pm$} \footnotesize{0.84}                     \\                               
SOP 50k *                     & 68.44 & 66.3  & 58.52 & 32.84                                                                                         
\\ 
NCOD 50k                   & \textbf{74.86}
& \textbf{70.72}
& \textbf{62.94}
& \textbf{40.5}
\\
              & \footnotesize{$\pm$} \footnotesize{0.14} & \footnotesize{$\pm$} \footnotesize{0.11} & \footnotesize{$\pm$} \footnotesize{0.22} & \footnotesize{$\pm$} \footnotesize{1.21}                                                                                   
\\

\end{tabular}
\caption{CIFAR-100 test accuracy with Resnet 34 under varying symmetrical noise percentages. Mean and standard deviation are recorded from three runs, with best performances are in bold. * indicates the experiments we ran again using the official implementation (other results are from the papers). 
The number associated with the method name represents the number of samples used in training. }
\label{tab:simple100}
\end{table}

Results have proved that our approach is superior to other methods at all levels of noise for CIFAR-100, as shown in the comparison tables. To reproduce the results of the SOP method \cite{liu2022robust} we have used a reduced set of $45k$  training samples for CIFAR-100 as mentioned in their work.  (while the standard size for the training set for CIFAR-100 is $50k$).
The results have shown that as the number of training samples increases, the performance of our model improves, while we observed a drop in performance for the SOP method for some levels of noise.
We also achieve the state of art performance using pre-trained models for mini-web vision. 
Training InceptionResnetV2 from scratch with mini-web view and the hyperparameters specified in \cite{liu2022robust} we were not able to reproduce the results in the paper. Our model, instead, obtained better results for the same experimental settings, see \cref{tab:miniwebvison}.




\begin{table}
\centering
\begin{tabular}{l|l|l|l|l} 
\hline
                                                                      & \multicolumn{4}{c}{~ ~ ~ ~ ~ ~ ~ Method}                                                                                                                                 \\ 
\hline
Network                                                               & CE    & \begin{tabular}[c]{@{}l@{}}SOP\\paper\end{tabular} & \begin{tabular}[c]{@{}l@{}}SOP\\We ran\end{tabular} & \begin{tabular}[c]{@{}l@{}}NCOD~\\(ours)\end{tabular}  \\ 
InceptionResNetV2                                                     & -     & 76.6                                               & 76.5                                                & 78.44                                                  \\ 
\begin{tabular}[c]{@{}l@{}}Resnet50\\pretrained\end{tabular}          & 81.15 & -                                                  & 82.28                                               & \textbf{83.33}                                                  \\ 
\begin{tabular}[c]{@{}l@{}}InceptionResnetV2\\pretrained\end{tabular} & 81.06 & -                                                  & 82.36                                                  & \textbf{83.49}                                                  \\
\end{tabular}
\caption{Test accuracy for the mini web vision dataset }
\label{tab:miniwebvison}
\end{table}

Supplementary materials, including code, experimental settings, and extra figures, are available at: \\
\url{https://github.com/wanifarooq/NCOD}

\section*{Conclusions and Future Directions}

We presented NCOD, a technique designed for training deep neural networks in the presence of noisy labels. Our method effectively addresses noise, outperforming other approaches as shown in our experiments. By using on the inherent similarities among samples within a class and that networks prioritize learning from clean data, we utilize class embeddings and outlier discounting to enhance accuracy and resilience to noise across diverse datasets.
In future work, we aim to refine our method by investigating alternative discount functions beyond just linear discounting and exploring varied loss functions and learning strategies for the parameter $u$. Concerning the influence of $u$, we also intend to incorporate this idea into other strategies from literature, adapting the $\mathcal{L}_2$ loss in Eq. \ref{eq:loss} to losses proposed by different methods.



\bibliographystyle{abbrv}
\bibliography{bibliography}
\onecolumn
\appendix
\begin{center}
\textbf{\Large Supplementary Materials}
\end{center}
\section{Training details for NCOD and NCOD+}

\subsection{Definition of label Noise}
In this paper, we employed two distinct categories of datasets: those containing real label noise and those with synthetic label noise. In the case of the CIFAR-100 dataset, we created artificial noise by implementing the techniques detailed in the study conducted by \cite{liu2022robust}. Conversely, the mini Webvision dataset is a real noisy dataset. This strategy enables us to assess the efficacy of our approach on datasets featuring both synthetic and real label noise, thereby offering a comprehensive insight into its performance across diverse scenarios.

\subsection{NCOD+}
To use NCOD+ for the ensemble network architecture experiments, we have also employed two additional regularization terms as used in \cite{liu2022robust}. The definition of these two regularizations are taken directly from their work and are used to improve the consistency and class balance of the network's predictions. These regularization terms are the consistency regularizer $L_C$ and class-balance regularizer $L_B$. 
Both of these terms are defined using the Kullback-Leibler divergence and help to improve the accuracy of the network's predictions by encouraging consistency and preventing the network from assigning all data points to the same class.
The final loss function for NCOD+ is constructed by combining these regularization terms with our original loss function.

\subsection{Network structures and hyperparameters} \label{Appendix:structures_hyperparameters} In our experiments, we utilized the Torch library version 1.12.1 and used specific hyperparameters and network types for NCOD and NCOD+ on different datasets. These details are provided in \cref{tab:hyperparameters}. It is worth noting that for particular experiments, retraining the last saved model for a small number of additional epochs, e.g., from 5 to 10, improved the performance of the model. This highlights the effectiveness of our method even when applied to pre-trained models. Also, to avoid computational overhead, we calculated the mean representation of each class only once for each epoch, particularly at the beginning of each epoch.

\subsection{Experimental Settings}
In our experiments on CIFAR-100, we use simple data augmentations including random crop and horizontal flip, following the same approach used in previous works (\cite{patrini2017making};\cite{liu2020early}). To improve the performance of our method, NCOD+, we also use unsupervised data augmentation techniques as described in (\cite{augmentation}) to create additional views of the data for consistency training.
For the  mini WebVision, we first resize the images to 256x256, and then perform a random crop to 227x227, along with a random horizontal flip. All images are standardized by means and variances to ensure consistency during the experiments.
In our experiments, we use the Stochastic Gradient Descent (SGD) optimizer without weight decay for parameter U. We keep all the hyperparameters fixed for different levels of label noise to ensure fair comparison across experiments. To perform a fair comparison; we use the same settings of hyperparameters and architectures for both NCOD and NCOD+. Table \ref{tab:hyperparameters} provides a  detailed description of the used hyperparameters. 

\begin{table}[H]
\centering
\resizebox{\columnwidth}{!}{%
\begin{tabular}{|l|l|l|l|l|l|} 
\hline
                  & \multicolumn{3}{c|}{CIFAR-100}                                                                                                                                                               & \multicolumn{2}{c|}{mini webvision}                                                                     \\ 
\hline
Architecture      & ResNet34                                                       & PreActResnet18                                               & PreActResnet34                                               & InceptionResNetV2 & \begin{tabular}[c]{@{}l@{}}InceptionResNetV2,\\ResNet-50 \\both(pretrained)\end{tabular}  \\ 
\hline
batch size        & 128                                                            & 128                                                          & 128                                                          &   32                &   256                                                                                  \\ 
\hline
learning rate     & 0.02                                                           & 0.02                                                         & 0.02                                                         &  0.02                 &  0.1                                                                                   \\ 
\hline
lr~ decay         & \begin{tabular}[c]{@{}l@{}}[80,120]\\multistep\\gamma =0.1\end{tabular} & \begin{tabular}[c]{@{}l@{}}Cosine \\Annealing \end{tabular}                                     & \begin{tabular}[c]{@{}l@{}}Cosine \\Annealing \end{tabular}                                           &  \begin{tabular}[c]{@{}l@{}}[50]\\multistep\\gamma =0.1\end{tabular}                 &  \begin{tabular}[c]{@{}l@{}}init 0.1\\LambdaLr\\warmup =5\end{tabular}                                                                                   \\ 
\hline
weight decay      & $5*10^{-4} $                                                     & $5*10^{-4}$                                                    & $5*10^{-4} $                                                   & $5*10^{-4} $                &  0.001                                                                                   \\ 
\hline
\begin{tabular}[c]{@{}l@{}}training \\epochs \end{tabular}   & $150$                                                             & $300$                                                          & $300$                                                         & $100$                  & $25$                                                                                    \\ 
\hline
\begin{tabular}[c]{@{}l@{}}training \\examples \end{tabular}  & 50 k                                                           & $50$k                                                          & $50$k                                                          &  $66k$                 &   $66k$                                                                                  \\ 
\hline
lr for u          & sym =$0.1$                                                       & \begin{tabular}[c]{@{}l@{}}sym = $0.1$\\Asym =$0.3$\end{tabular} & \begin{tabular}[c]{@{}l@{}}sym = $0.1$\\Asym =$0.3$\end{tabular} & $0.3$                  & $0.3$                                                                                    \\ 
\hline
wd for u          & $1e^{-8} $                                                       & $1e^{-8}$                                                      & $1e^{-8}$                                                      & $1e^{-8}$              &  $1e^{-8}$                                                                                   \\ 
\hline
init. std. for u        & $1e^{-9}$                                                       & $1e^{-9}$                                                      & $1e^{-9}$                                                      & $1e^{-9}$              &  $1e^{-9}$                                                                                    \\ 
\hline
$\lambda^c$                & $0.0$                                                            & $0.9$                                                          & $0.9$                                                          & $0.0$               &   $0.0$                                                                                  \\ 
\hline
$\lambda^b$                & $0.0$                                                            & $0.1$                                                          & $0.1$                                                          & $0.0$               &  $0.0$                                                                                   \\
\hline
\end{tabular}}
\caption{The table shows the hyperparameters used for our experiments, which were kept consistent across all experiments to ensure a fair comparison with other methods, particularly SOP (\cite{liu2022robust}). This consistency in hyperparameters allows for an accurate comparison of the performance of our method with respect to other methods and helps to eliminate any bias that may be introduced by variations in the experimental setup. The table provides a detailed description of the settings that were used to conduct the experiments, which is useful for reproducing the results and for understanding the experimental conditions under which our method was evaluated. It's also worth noting that, in case we retrain from the previous best-saved model, the learning rate for 'u' is given as high as 3. This is done to fine-tune the model and improve its performance. Additionally, The values of $\lambda^c $ and $\lambda^b$ are the coefficients for consistency regularization and class balance regularization respectively, as used in the experiments done by \cite{liu2022robust}. These coefficients are used to balance the trade-off between consistency and class balance in the model, and their values are determined through experimentation. }
\label{tab:hyperparameters}
\end{table}
\subsection{Additional details}
\begin{figure*}
\centering
\subfigure[Noisy samples]{
\includegraphics[width=0.45\linewidth]{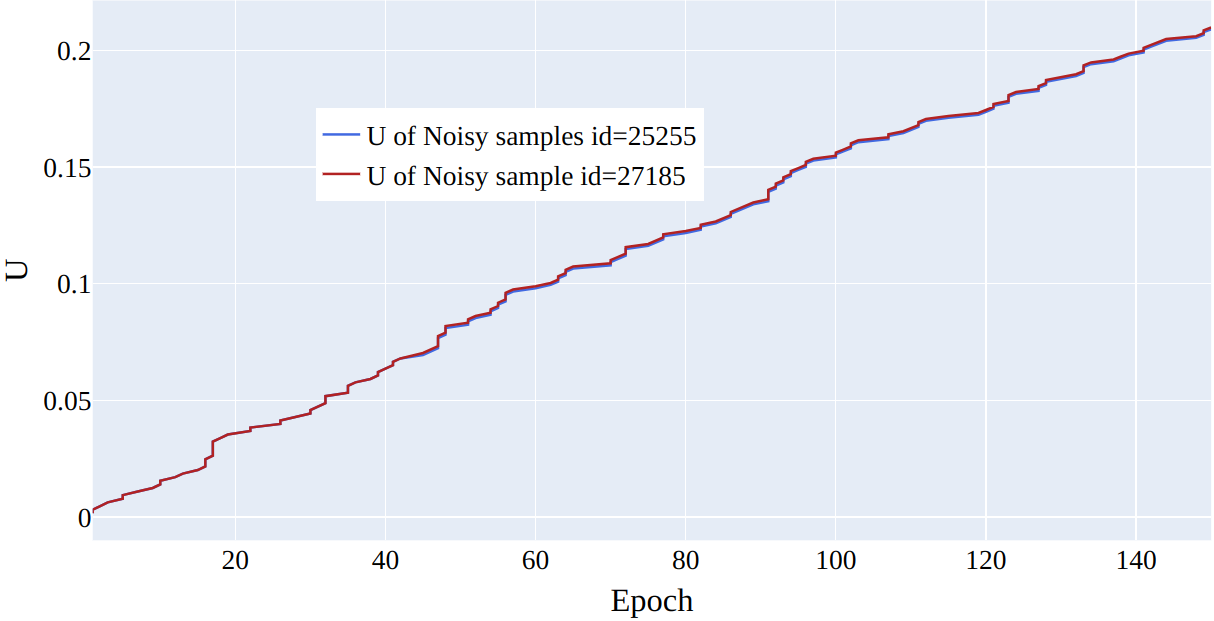}
}\quad
\subfigure[Pure samples]{
\includegraphics[width=0.45\linewidth]{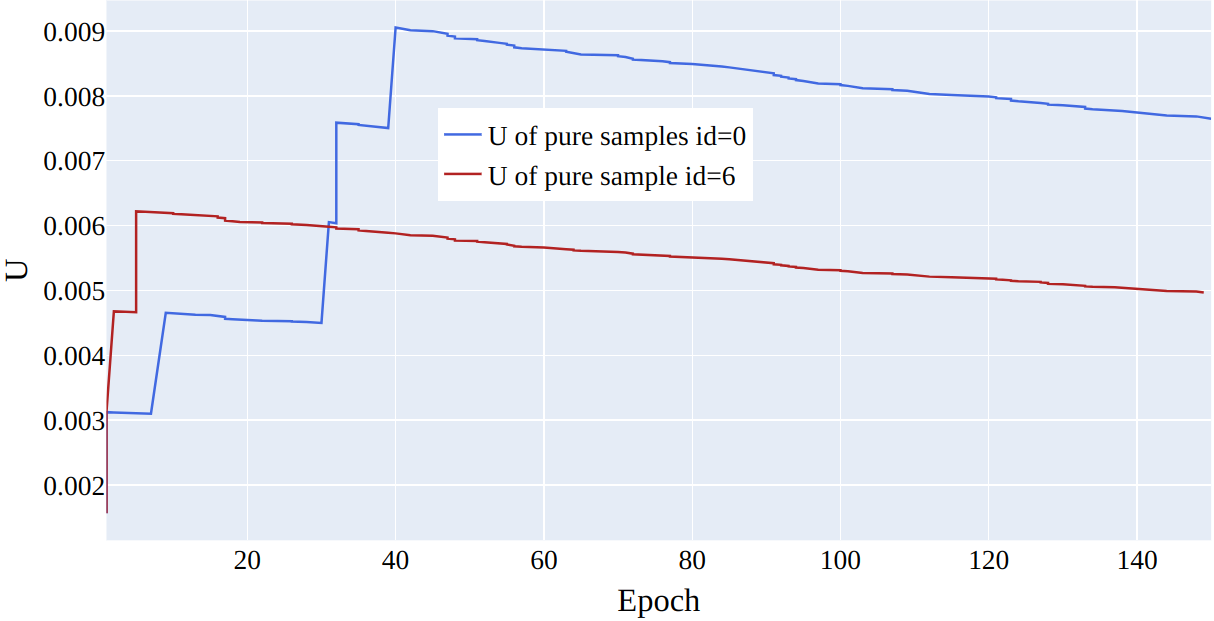}
}
\caption{
The figure depicts the learning of parameter $u$ for two noisy and two pure samples in $50\%$ symmetrical noise for CIFAR-100
}
\label{fig:sample_noise}
\end{figure*}

\begin{figure*}
\centering
\subfigure[Noisy samples]{
\includegraphics[width=0.45\linewidth]{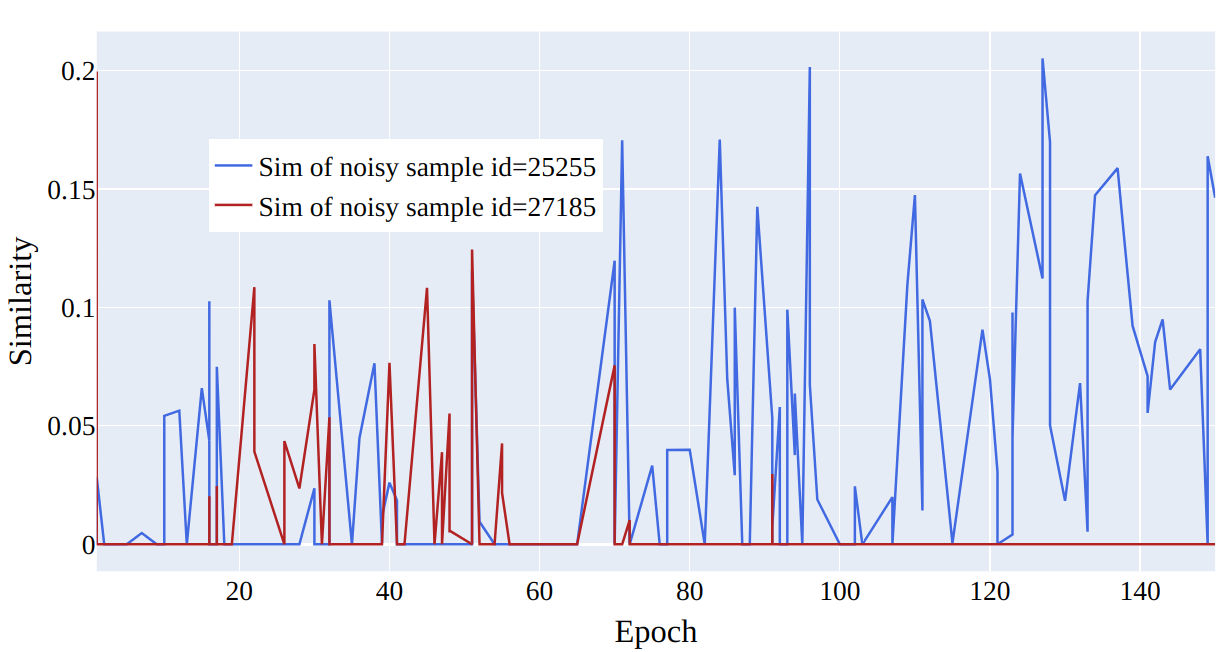}
}\quad
\subfigure[Pure samples]{
\includegraphics[width=0.45\linewidth]{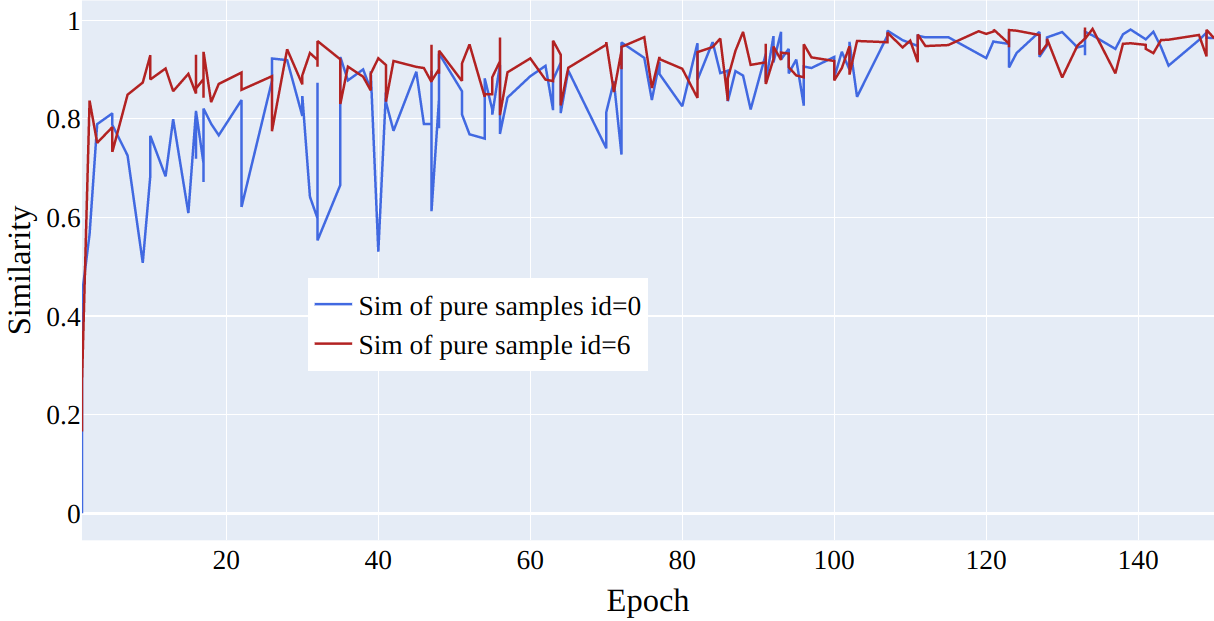}
}
\caption{The figure shows how similarity changes during training for two noisy and two pure samples in $50\%$ symmetrical noise for CIFAR-100}
\label{fig:sample_sim}
\end{figure*}

It is important to note that in Figure \ref{fig:Figure_u}, the average similarity of samples starts at around $0.20$ instead of $0$. This is because, in the first epoch, the class representative embeddings are initialized randomly, resulting in different similarity values and soft labels for each sample. This ensures that the noisy labeled data  has no initial advantage and can begin learning. As training continues, the similarity of pure samples increases and predictions improve, but at some point, it starts overfitting, and accuracy decreases. However, the inclusion of $u$ in our method prevents overfitting by removing the effect of noisy samples in the cross-entropy loss. As previously stated, in the first few epochs, even the similarity of pure samples is a bit low, causing $u$ to learn for them as well. But eventually, as their similarity increases, $u$ starts decreasing for pure samples. In the case of noisy samples, as their similarity drops, their predictions decrease, and $u$ increases to compensate  for the effect of this in $\mathcal{L}_1$ 
and to decrease the loss of $ \mathcal{L}_2 $.
The change in $u$ can be seen in Figure \ref{fig:sample_noise} for 2 noisy and 2 pure samples and their corresponding similarities can be seen in Figure \ref{fig:sample_sim}.

\section*{Further Analysis of Our Training Strategy}

For each class, $c$, the set of samples in the training dataset that have class $c$ as their label can be partitioned into $C$ subsets, where each subset is comprised of examples belonging to the same clean label.
To make learning from data possible, we must assume that each subset of samples with the corrupted label in $c$ should be smaller than the subset having the clean label of $c$. If that is not the case, then the observed label is independent of the data, and no learning is possible.
More formally, let $n_{c}$ be the number of samples for class $c$ in the training dataset. For each class $c$, we denote by $n_{c,c'}$ the number of samples labeled as $c$ whose correct label is $c'$ instead, and by $n_{c,c}$ the number of samples whose labels are correctly labeled as $c$. The above assumption can thus be expressed as $n_{c,c} \geq n_{c,c'} \; \forall \; c' \neq c, \; c' \in \{ 1, \dots, C\} $.

Our training procedure can be explained by considering three consecutive stages.

\paragraph{Initialization.} At the beginning of the learning phase, since the weights are initialized randomly, all the errors on the training samples are of comparable magnitude.
More formally, let $n$ be the number of samples of the training set; then, we can write the loss as
\begin{equation}
\begin{split}
&\sum_{i=1}^n  \mathcal{L}_{\theta}(f( \theta,x_i) + u_i\textcolor{black}{ \cdot y_i}, \tilde{y}_i  ) \\
&= \sum_{c = 1}^C  \sum_{j=1}^{n_c}\mathcal{L}_{\theta}(f( \theta,x_j) + u_j \textcolor{black}{ \cdot y_j}, \tilde{y}_j  )\\
 &= \sum_{c = 1}^C  \sum_{c'=1}^C \sum_{j=1}^{n_{c,c'} } \mathcal{L}_{\theta}(f( \theta,x_j) + u_j\textcolor{black}{ \cdot y_j}, \tilde{y}_j  )
\end{split}
\end{equation}

Given that initially, the representations for each sample are chosen randomly, the values of the loss $\mathcal{L}_{\theta}(f( \theta,x_j) + u_j\textcolor{black}{ \cdot y_j},\tilde{y}_j ) $ for each sample belonging to the class $c$, are all similar. For this reason, the major contribution to the final value of the loss is given by the largest subset of samples, which in this case is the subset of samples correctly labeled. To minimize the loss, thus, the best strategy consists of minimizing the loss on that subset of samples, and the gradient will point to the direction mostly aligned with their gradients. Therefore, the updated representations will be influenced mostly by the samples with the correct labels. 

\paragraph{Neighborhood Similarity and Avoiding Memorization.} The similarity between samples and their class embeddings is the primary factor contributing to the loss due to two phenomena. Firstly, correct samples are more likely to be close to the right class embedding, and secondly, their subset carries more weight in determining the average location in the embedding space. Higher similarity with the class embedding leads to a larger contribution to the loss. However, the model tends to overfit on noisy labels, which is addressed by introducing the outlier discounting parameter $u$ in our loss, reducing the weight of samples with high $u$ values.

\begin{figure}[ht]
    \centering
    \includegraphics[width=0.9\linewidth]{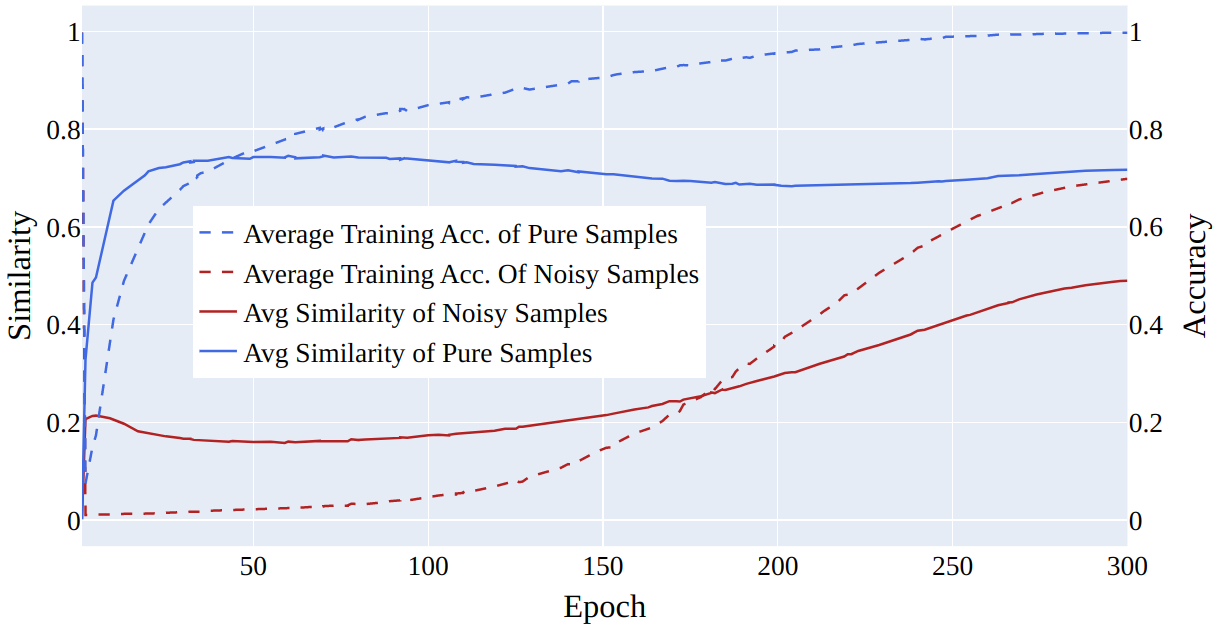}
    \caption{ Shows the model initially learns for pure samples which correspond to higher similarities and after approx. 100 epochs it significantly starts learning for smaller insignificant similarities corresponding to noisy samples.
   We obtained the plot using $50\%$ symmetrical noise.
    }
    \label{fig:avg_sim_comp}
\end{figure}

\begin{figure}
    \centering
    \includegraphics[width=0.9\textwidth]{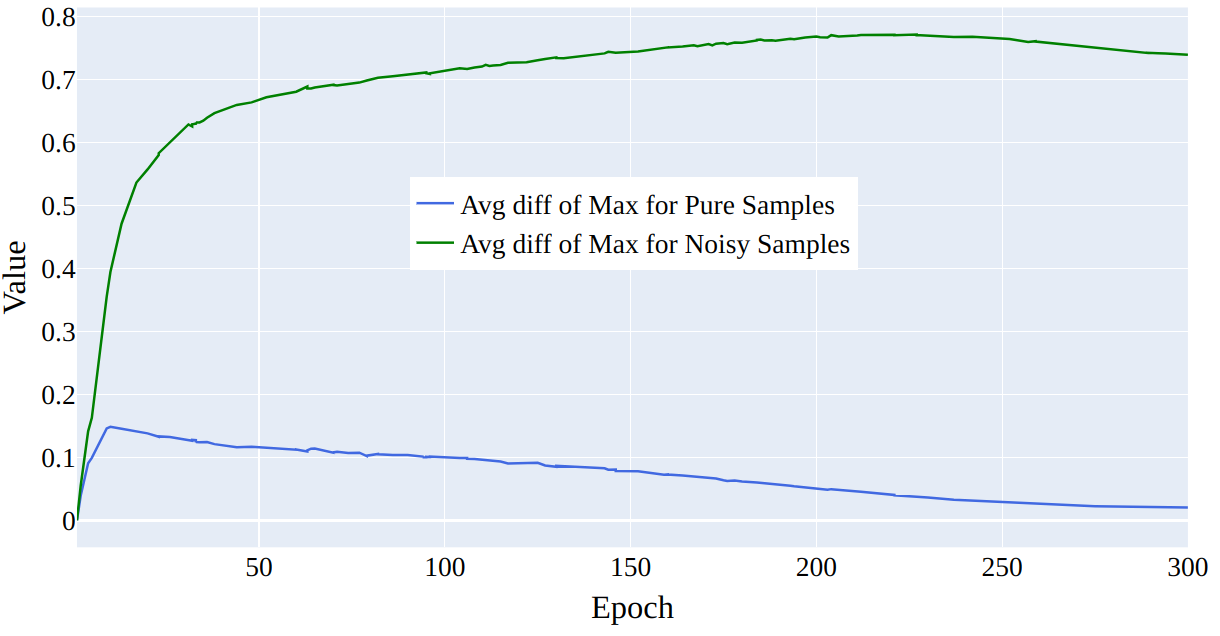}
    \caption{Average difference between the  probability of the predicted class  and the probability assigned to the label in the dataset for both noisy and pure samples.}
    \label{fig:max_diff}
\end{figure}

In \cref{fig:max_diff} we plot the computed average of the difference between the probability assigned to the predicted class and the probability assigned to the true label for both noisy and pure samples. More in detail, given a sample $(x_i,y_i)$ we are considering a network output represented by the vector $f(\theta, x_i)$ of $C$ components after applying softmax. Here, $y_i$ represents the label the sample has in the dataset, for the sample, denoted by $c$. We can denote the individual components of the vector as $f(\theta, x_i)_j$ for $j = 1,..., C$.
 
The quantity that we are plotting is given by the difference between the maximum value of 
$f(\theta,x_i)_j$ across all $j$ and the component $f(\theta,x_i)_c$ corresponding to the label $y_i$. In other words, we are computing the expression:
$\max{({j=1, ..., C} f(\theta,x_i)_j)} - f(\theta,x_i)_c$

\section{Proofs} \label{appendix:study_series}
\begin{proposition*} [\ref{increasing_of_u}]
Let $(x_i, y_i)$ be a sample, $\theta^t$ be the parameters of the network at epoch $t$, and $u^t_i$ be the parameter for outlier discounting relative to sample $i$ at epoch $t$. Let $ \hat{c}^t_i$ be the prediction of the network at time $t$, and $y_i$ be the class of sample $i$.
Suppose $\hat{c}^t_i = y_i$, then $u^{t+1}_i < u^{t}_i$ when $u^{t}_i \neq 0$, and $u^{t+1}_i =0 $ otherwise. 
When $\hat{c}^t_i \neq y_i$, $u^{t+1}_i \geq u^{t}_i$.
Moreover $u_i^t<1$ for $t \in \mathbb{N}$. 
\end{proposition*}
\begin{proof}

We write $\hat{c}_i(t)$, to emphasize that also the class predicted by the network depends on the epoch since the network parameters are updated accordingly \cref{eq:joint_update}.
 The update rule for $u $ will be
\begin{equation}
    \begin{split}
        u^{t+1}_i &= u^{t}_i - \beta \; \partial_{u_i} \mathcal{L}_2   \\
         &= u^t_i - 2\frac{\beta}{C} ( \delta_{c_i,\hat{c}_i(t)} -1 + u^t_i)
    \end{split}
\end{equation}
$\delta_{c_i,\hat{c}_i}$ is $1$ if the label predicted by the network coincides with the label of the sample in the dataset while it is $0$ otherwise.
So the update becomes:
\begin{equation*}
 u^{t+1}_i  = \begin{cases}
 u^t_i(1 - 2\frac{\beta}{C}) \text{ if } \hat{c}_i(t)= c_i  \\
u^t_i(1 - 2\frac{\beta}{C}) + 2\frac{\beta}{C} \text{ if }  \hat{c}_i(t) \neq c_i 
 \end{cases}
\end{equation*}
Notice that in our setting $\beta$ is the learning rate used to learn the outlier discounting and $C$ is the number of classes, so $2\frac{\beta}{C} <1$.
From the equation above it follows that if $u^0_i = 0$, $$u^{t+1}_i = 2 \frac{\beta}{C}\sum_{k=0}^t \left( 1- 2 \frac{\beta}{C} \right)^{k} \left(1 - \delta_{\hat{c}_i(t-k),c_i} \right) $$
From this writing, we can see that the maximum of the sum is reached when predictions are always wrong. In this case, the sum becomes a geometric sum with a ratio smaller than one and we obtain $\lim_{t \to \infty}u_i^{t} = 1$.
We can also notice that if the classes coincide $ u^{t}$ is multiplied by a value less than one, so it either decreases in magnitude or remain at zero if its initial value was zero.
If the prediction is incorrect, since $u_i^{t}<1$, $u^{t+1}$ increases.  Indeed  $ u^t_i(1 - 2\frac{\beta}{C}) + 2\frac{\beta}{C} > u^t_i$ if and only in $u^t_i<1$.
\end{proof}

\subsection{Proof Remark \ref{remark_position_of_u}} \label{appendix:proof_remark}
\begin{proof}
    Let $c$ be a constant, $0< c <1$, we consider the function $f(x) = \log(x+c) - (1-c)\log(x)$.  For $x>0$, $f$ is continuous and derivable and 
    $f'(x) = \frac{1}{x+c} - \frac{1-c}{x}$. 
    Studying the sign of the derivative we can observe that $f'(x) \geq 0$ if $x \geq 1-c$. It follows that $f$ has a minimum in  $x = 1-c$ and since $1-c <1$,  $f(1-c) = - (1-c)\log(1-c)>0 $, namely the function $f$ is always positive, and the desired inequality is valid.
\end{proof}

\section{Latent Representations}
\label{appendix:Latent_Representations}

\begin{figure}[h!]
\centering
\subfigure[Epoch 1]{
\includegraphics[width=0.40\linewidth]{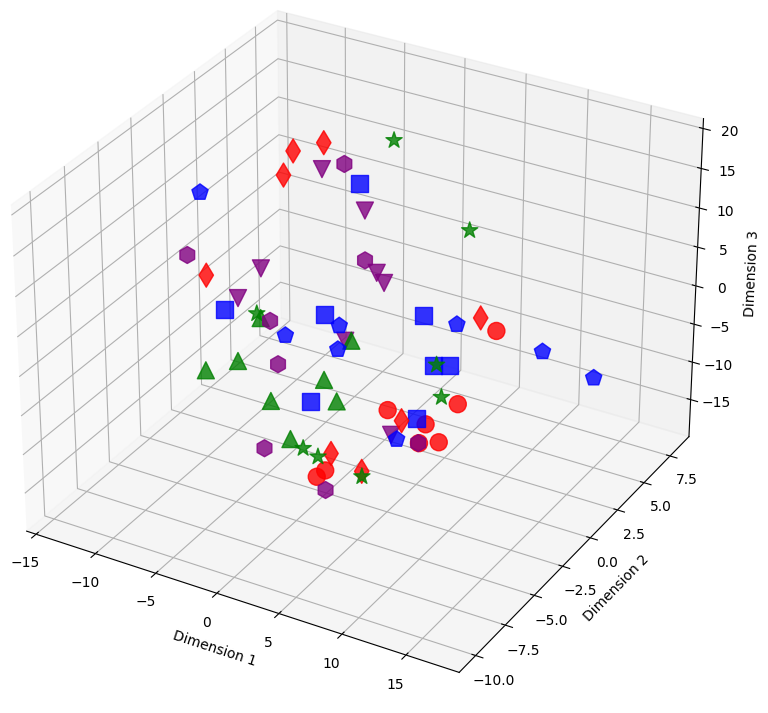}
}\quad
\subfigure[Epoch 14]{
\includegraphics[width=0.40\linewidth]{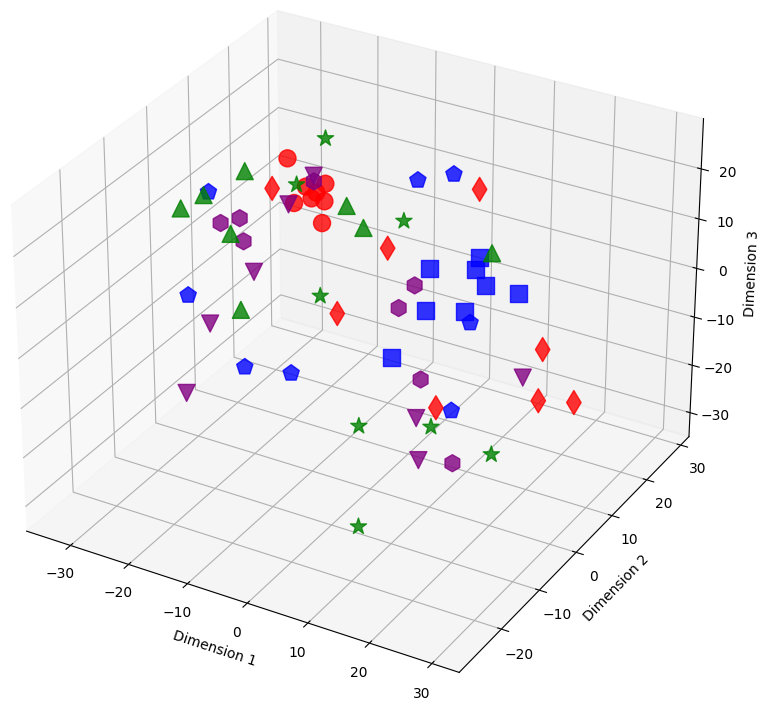}
}\quad
\label{fig:Training_our_20_e}
\caption{  The figure illustrates sample embeddings of four classes from CIFAR-100 with 20\% symmetrical noise at different training epochs using  our loss function (NCOD). Colors represent classes, and shapes distinguish noisy and pure samples: Blue (square: pure, pentagon: noisy), Red (circle: pure, diamond: noisy), Green (triangle-up: pure, star: noisy), and Purple (triangle-down: pure, hexagon: noisy).
}
\end{figure}

\begin{figure}
\centering
\subfigure[Epoch 1. CE Loss]{
\includegraphics[width=0.40\linewidth]{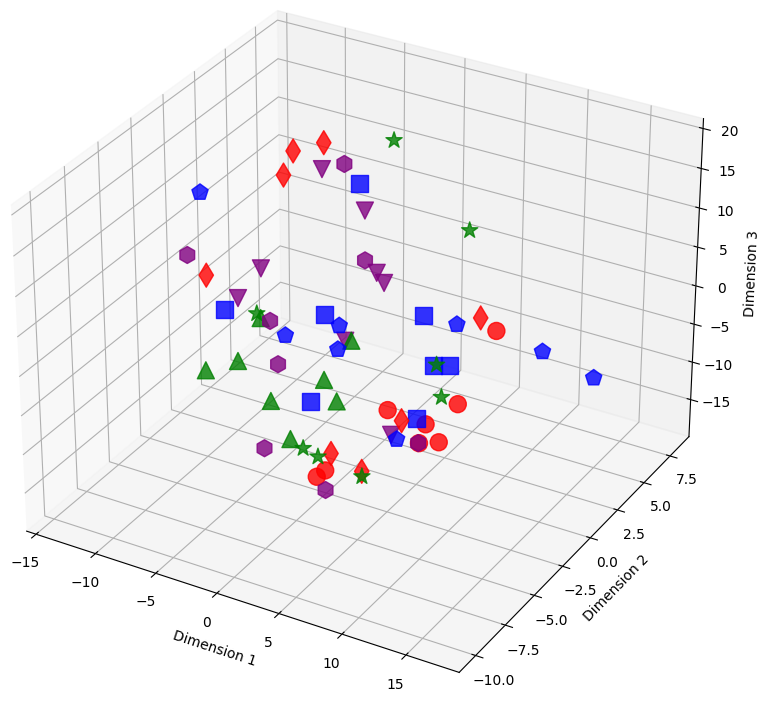}
}\quad
\subfigure[Epoch 14. CE Loss]{
\includegraphics[width=0.40\linewidth]{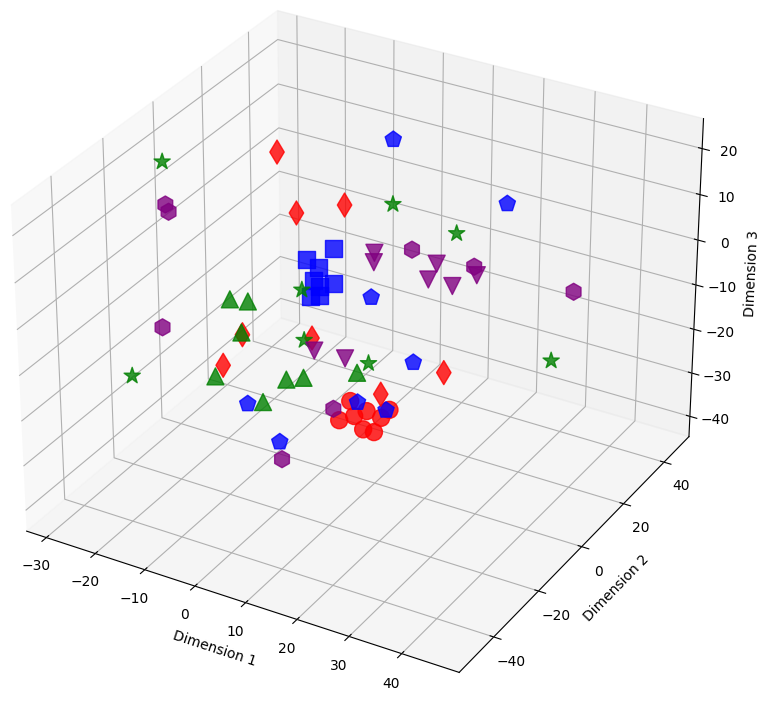}
}\quad
\subfigure[Epoch 96. CE Loss]{
\includegraphics[width=0.40\linewidth]{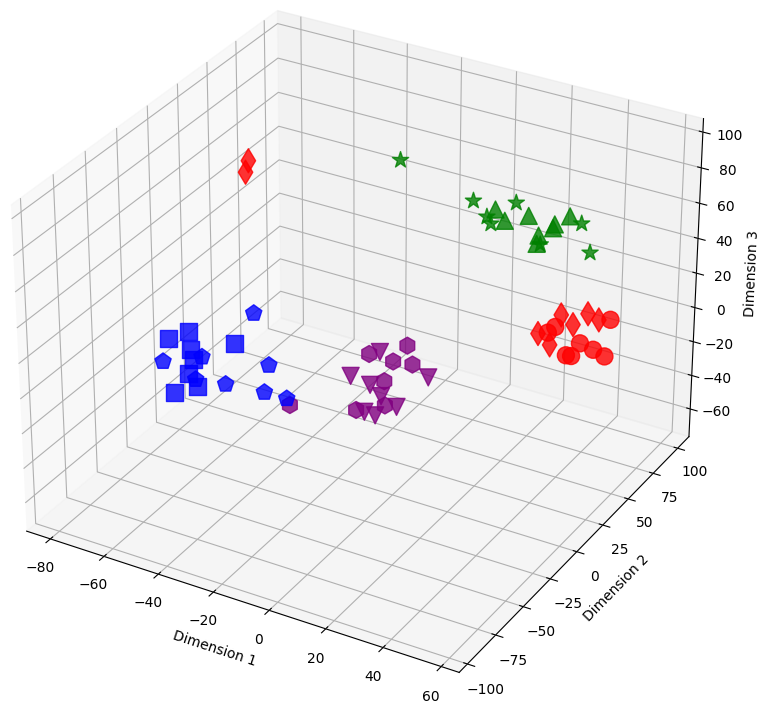}
}\quad
\subfigure[Epoch 96. NCOD Loss]{
\includegraphics[width=0.40\linewidth]{Figures/Embeddings/Our_loss_20_percentage_CIFAR100/96.png}
}\quad
 \caption{ Sample embeddings of four classes from CIFAR-100 with 20\% symmetrical noise.
 Colors represent classes, and shapes distinguish noisy and pure samples: Blue (square: pure, pentagon: noisy), Red (circle: pure, diamond: noisy), Green (triangle-up: pure, star: noisy), and Purple (triangle-down: pure, hexagon: noisy).
\label{fig:Enlarged_version_figure1_main}
 }
\end{figure}

The 3D visualizations in \cref{fig:Training_our_20_e} show the evolution of latent representations in the penultimate layer of the model during the training process.
The figures in this section show the shifts of these representations for training samples selected from the CIFAR-100 dataset with 20\% symmetric noise for different epochs of training using our loss, NCOD.  In order to make the plots easier to understand we choose to consider only samples belonging to four classes, in particular classes 0,1,2,3 from CIFAR 100 that correspond to 
 ``Apple'', ``Aquarium Fish'', ``Baby'' and ``Bear'' respectively.

Different colors represent classes. For each color, there are two different shapes that distinguish noisy and pure samples: Blue (square: pure, pentagon: noisy), Red (circle: pure, diamond: noisy), Green (triangle-up: pure, star: noisy), and Purple (triangle-down: pure, hexagon: noisy).

In this section, we included the missing figure showing the latent space for epochs 1 and 14 when using our loss function \cref{fig:Training_our_20_e}. We also add an enlarged version 
\cref{fig:Training_bce_and_ours_20}, that is \cref{fig:Enlarged_version_figure1_main}. 
As can be seen from the figures, in the first epoch, for both losses, the arrangement of samples in latent space is causal. 
Later, around epoch 14 we begin to see clusters related to the corrected samples (also for both losses) while the noisy samples are scattered in space.
Later in training, we can see completely different behavior between our loss and cross entropy, indeed at epoch 96 for cross entropy loss four clusters related to the four classes were formed, and the noisy examples are also part of these clusters, see subfigure (c) in \cref{fig:Enlarged_version_figure1_main}. Using our loss instead the clusters relative to the four classes are still visible but it can be observed that noisy samples stay at a distance from clusters related to classes, subfigure (d) in \cref{fig:Enlarged_version_figure1_main}.

We utilize t-SNE to reduce the 512 dimensional embeddings to a 3D dimensional embedding. Despite the reduction in dimensionality the formation of clusters and the different behavior of latent representations of noisy and pure samples is still visible.

\section{Distribution of the distance from Class Centroids} \label{appendix:distributions}

\begin{figure}[h!]
\centering

\subfigure[Epoch 1. Cross Entropy Loss]{\label{fig:Training_ce_20_d_epoch_1}
\includegraphics[width=0.40\linewidth]{Figures1/all1.png}
}\quad
\subfigure[Epoch 18. Cross Entropy Loss]{
\label{fig:Training_ce_20_d_epoch_18}
\includegraphics[width=0.40\linewidth]{Figures1/all16.png}
}\quad

\subfigure[Epoch 96. Cross Entropy Loss]{\label{fig:Training_ce_20_d_epoch_96}
\includegraphics[width=0.40\linewidth]{Figures1/all96.png}
}\quad
\subfigure[Epoch 124. Cross Entropy Loss]{\label{fig:Training_ce_20_d_epoch_124}
\includegraphics[width=0.40\linewidth]{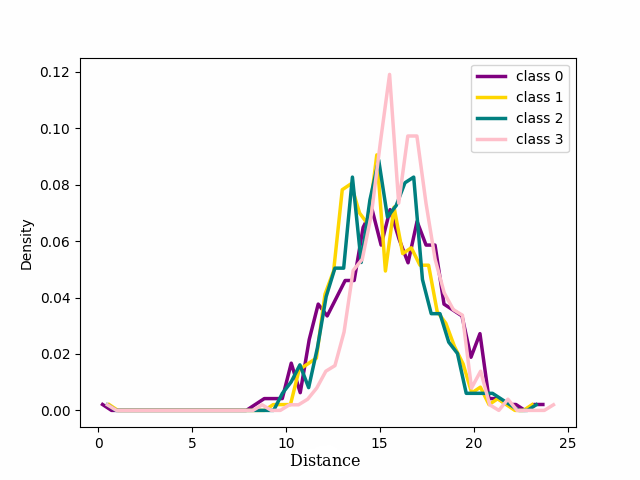}
}\quad
\subfigure[Epoch 1. NCOD Loss]{\label{fig:Training_our_20_d_epoch_1}
\includegraphics[width=0.40\linewidth]
{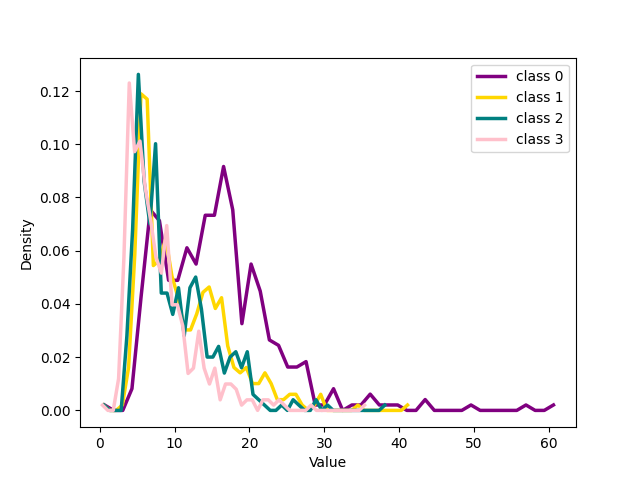}
}\quad
\subfigure[Epoch 18. NCOD Loss]{
\label{fig:Training_our_20_d_epoch_18}
\includegraphics[width=0.40\linewidth]{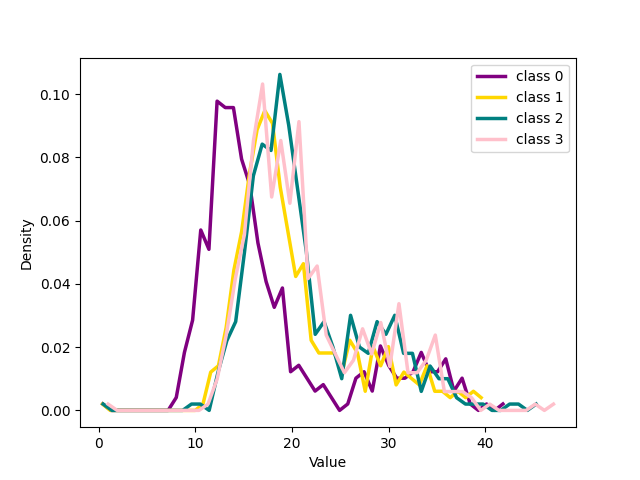}
}\quad
\subfigure[Epoch 96. NCOD loss]{\label{fig:Training_our_20_d_epoch_96_appendix}
\includegraphics[width=0.40\linewidth]{Figures/Distributions/Cifar100/Ours_20_perc/all96.png}
}\quad
\subfigure[Epoch 124. NCOD loss]{\label{fig:Training_our_20_d_epoch_124}
\includegraphics[width=0.40\linewidth]{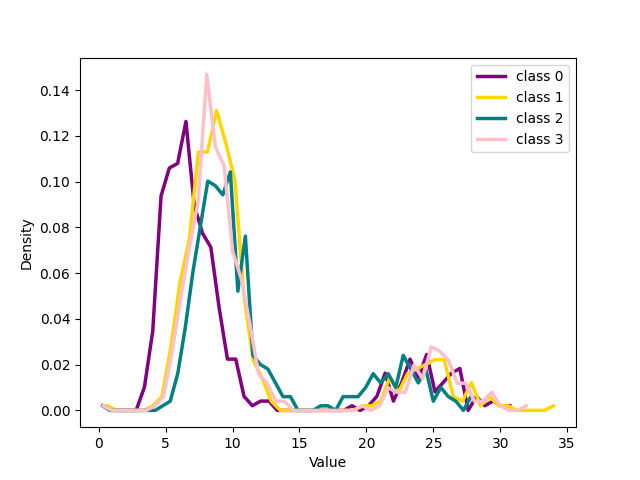}
}\quad
\label{fig:Training_our_20_d}
\caption{  The figure illustrates the distribution of four classes from CIFAR-100 with 20\% symmetrical noise at different training epochs using our loss function and cross-entropy loss function.
}
\end{figure}

The results in the previous section show a qualitative idea of what happens in training using the two losses. To get a quantitative measure Of the behavior of noisy and clean samples with respect to the formation of clusters in latent space  we chose to measure the distribution of samples around the centres of the clusters, the seeds, as explained in the main paper.
The \cref{fig:Training_our_20_d} shows the distribution of the distance in latent space between the seed of each of the four classes chosen for \cref{fig:Training_bce_and_ours_20} and the other samples belonging to those classes for CIFAR 100 dataset with 20\% symmetrical noise.

\begin{figure}[h!]
\centering
\subfigure[Epoch 18. CE loss. Noisy Samples]{
\includegraphics[width=0.40\linewidth]{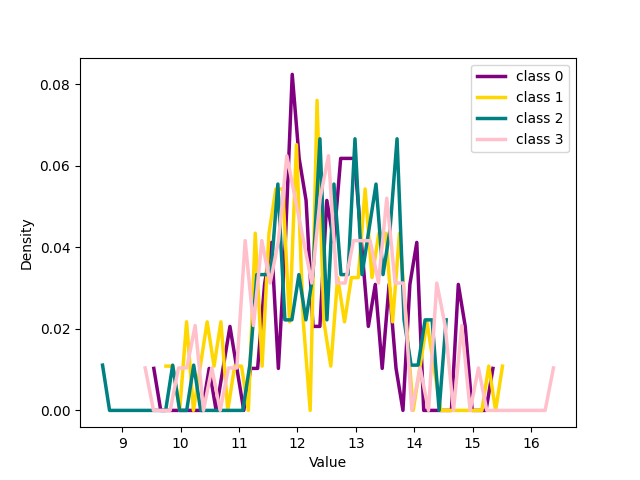}
}\quad 
\subfigure[Epoch 18. CE loss. Clean Samples]{
\includegraphics[width=0.40\linewidth]{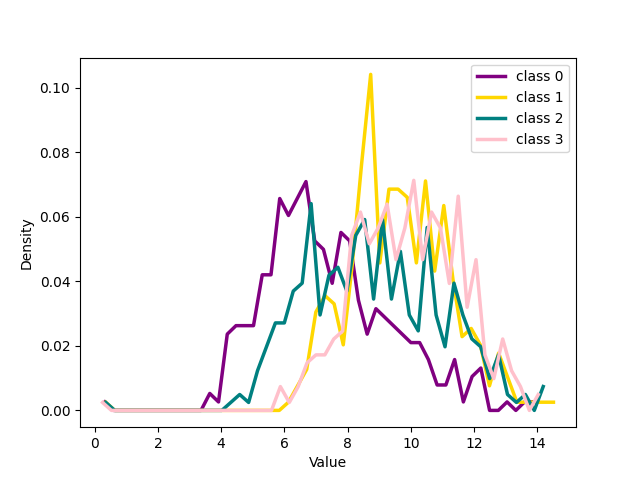}
} \quad 
\subfigure[Epoch 96. CE loss. Noisy Samples]{
\includegraphics[width=0.40\linewidth]{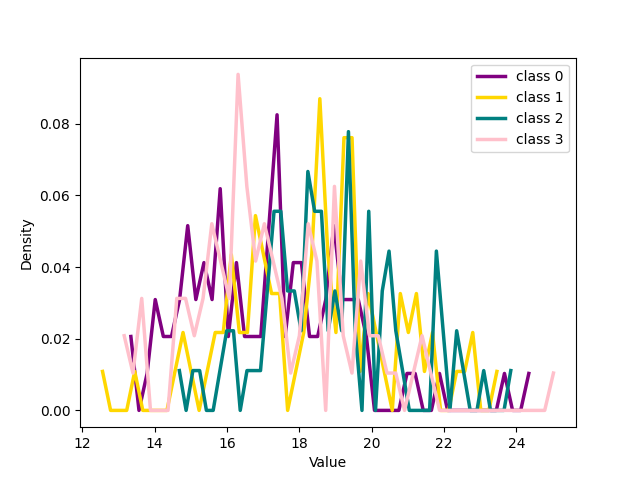}
} \quad 
\subfigure[Epoch 96. CE loss. Clean Samples]{
\includegraphics[width=0.40\linewidth] {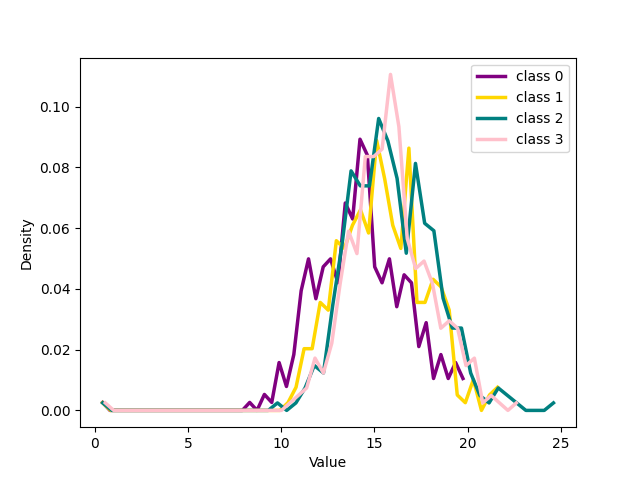}
} \quad 
\subfigure[Epoch 18. NCOD loss. Noisy Samples]{
\includegraphics[width=0.40\linewidth]{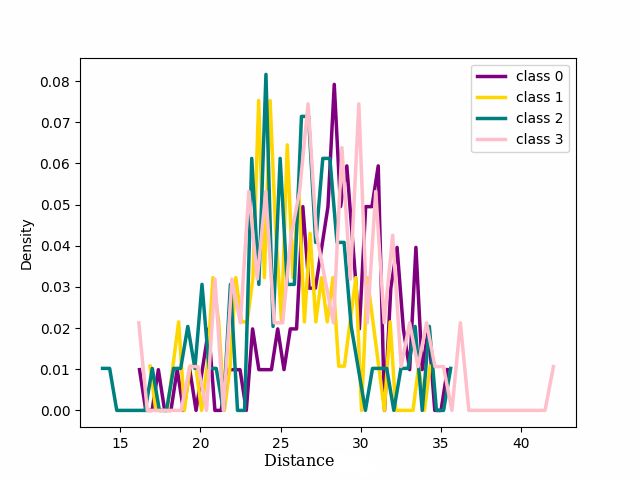}
 }\quad 
\subfigure[Epoch 18. NCOD loss. Clean Samples]{
\includegraphics[width=0.40\linewidth]{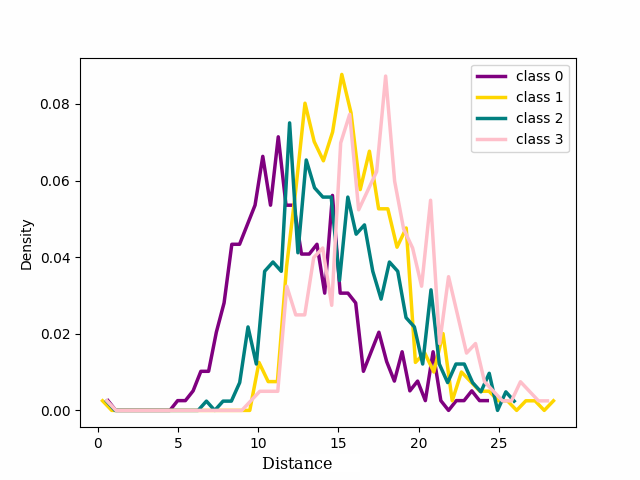}
} \quad 
\subfigure[Epoch 96. NCOD Loss. Noisy Samples]{
\includegraphics[width=0.40\linewidth]{Figures/Distributions/Cifar100/Noisy_and_Clean_separated/NCOD/noise96.png}
} \quad 
\subfigure[Epoch 96. NCOD loss. Clean Samples]{
\includegraphics[width=0.40\linewidth]{Figures/Distributions/Cifar100/Noisy_and_Clean_separated/NCOD/pure96.png}
} \quad 
\caption{\label{fig:distribution_noisy_and_clean_separated_cross_entropy} The figure illustrates the distribution of the four classes from CIFAR 100 with 20\% symmetrical noise for noisy and clean samples respectively at different training epochs using Cross Entropy loss or NCOD loss function.
}
\end{figure}

We can see that for both our loss and cross entropy loss at the beginning of training the distribution of distances at epoch 1 is unimodal, see \cref{fig:Training_bce_and_ours_20} subfigure (a) and (e), while at epoch 18, for both, we can see that some classes have a bimodal distribution, we have two peaks in the data, indicating the presence of two different groups, subfigure (b) and (f). 
For our loss, \cref{fig:Training_bce_and_ours_20} subfigure (f) the first peak is centered respectively around 15-18 depending on the classes, while the second peaks for all the classes are centred around 25-30, depending on the classes.
Plotting  the same plot only for noisy and clean sample separated shows exactly that for noisy samples is unimodal the peak is 
exactly around 25-30 while the distribution for the pure samples is unimodal with a peak centered around 15. This shows that the two groups of the bimodal distribution are exactly noisy and clean samples.
This can be observed in more detail in \cref{fig:distribution_noisy_and_clean_separated_cross_entropy} which shows the distributions of distances from seeds separately for noisy and clean samples.

\begin{figure}[h!]
\centering
\subfigure[Epoch 0. Cross entropy Loss]{
\includegraphics[width=0.40\linewidth]{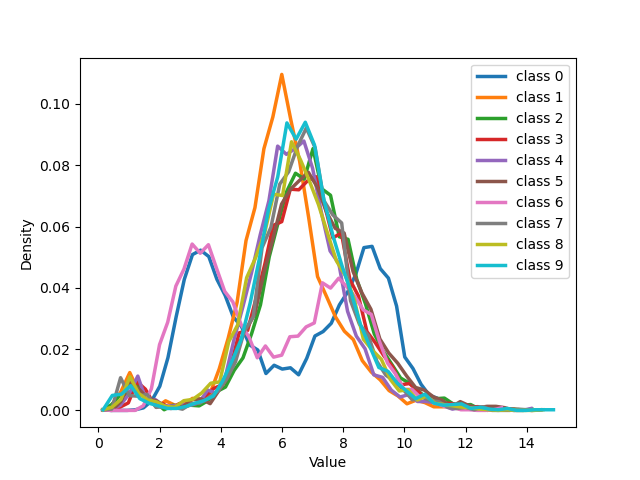}
}\quad
\subfigure[Epoch 1. Cross entropy Loss]{
\includegraphics[width=0.40\linewidth]{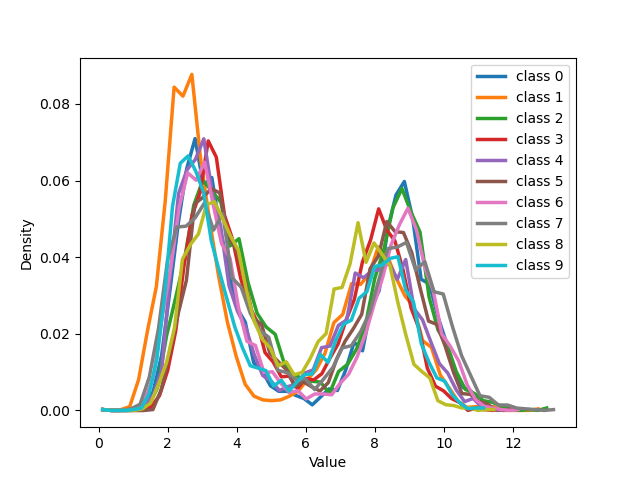}
}\quad
\subfigure[Epoch 10. Cross entropy Loss]{
\includegraphics[width=0.40\linewidth]{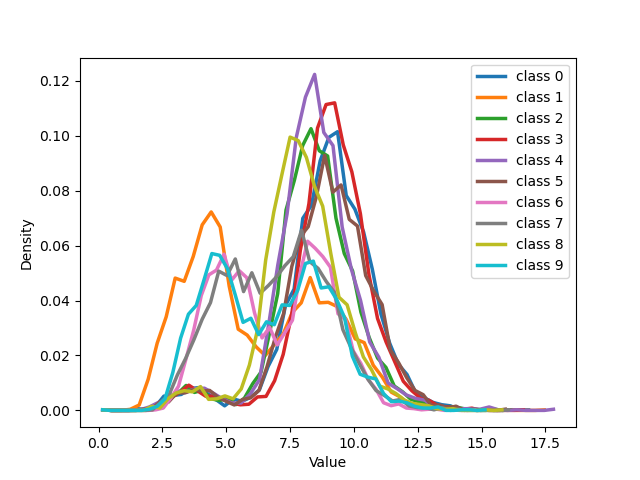}
}\quad
\subfigure[Epoch 20. Cross entropy Loss]{
\includegraphics[width=0.40\linewidth]{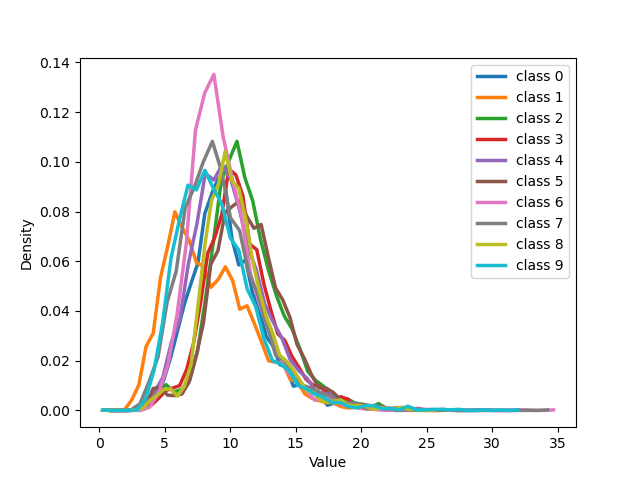}
}\quad
\subfigure[Epoch 0. NCOD loss]{
\includegraphics[width=0.40\linewidth]{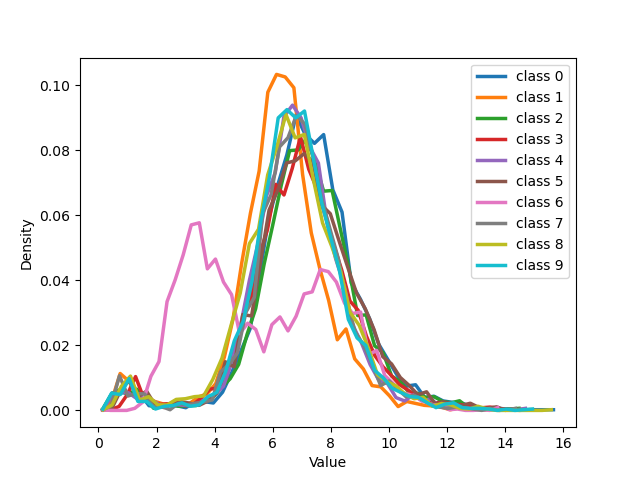}
}\quad
\subfigure[Epoch 1. NCOD loss]{
\includegraphics[width=0.40\linewidth]{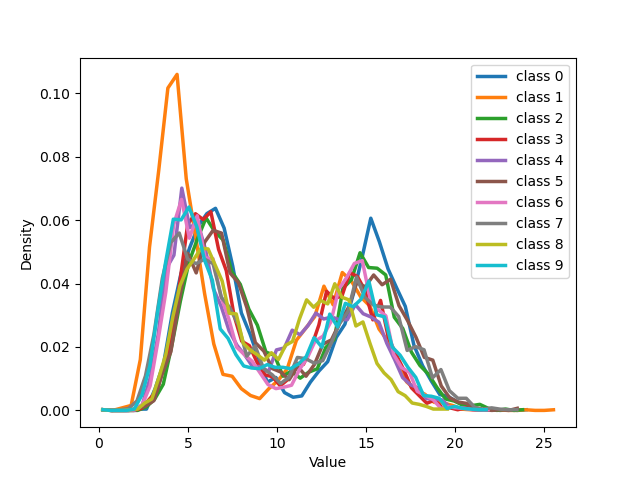}
}\quad
\subfigure[Epoch 10. NCOD loss]{
\includegraphics[width=0.40\linewidth]{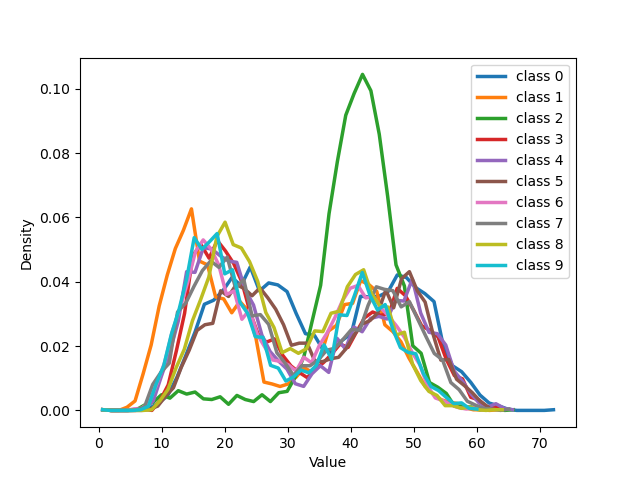}
}\quad
\subfigure[Epoch 20. NCOD loss]{
\includegraphics[width=0.40\linewidth]{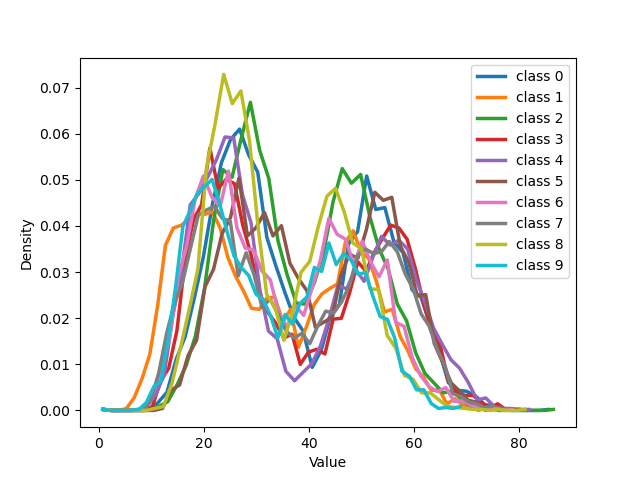}
}
\caption{\label{fig:Training_mbc_20_d}  The figure illustrates the distribution of all classes from MNIST with 50\% symmetrical noise at different training epochs using Cross Entropy loss and NCOD loss.
}
\end{figure}

Examining \cref{fig:Training_mbc_20_d} and delving into the insightful distribution analyses of sample embeddings conducted on the CIFAR-100 dataset, an intriguing parallel emerges with the results seen in the MNIST dataset. Notably, both the cross-entropy (CE) and our novel NCOD loss function initially exhibit analogous behavior, commencing with an unimodal distribution as captured in (a) and (e) of \cref{fig:Training_mbc_20_d}, then transitioning through training to a bimodal distribution as depicted in (b) and (f) of \cref{fig:Training_mbc_20_d}. However, as training progresses further the distribution of embeddings for the cross-entropy method begins to shift back towards an unimodal state, shown in (c), the embeddings under NCOD retain their bimodal nature, as demonstrated in (g). This captivating pattern persists, culminating in the eventual complete unimodality of distribution for CE, as seen in (d), juxtaposed against the unwavering bimodality of NCOD, magnificently depicted in (h). It's important to underscore the significance of this behavior. A bimodal distribution signifies a perceptive distinction between pure and noisy samples, while an unimodal distribution lacks this discriminatory power. This revelation adds an extra layer of brilliance to our findings.
\begin{figure}[h!] 
\centering
\subfigure[MNIST with 50\% symmetric noise]{\label{fig:train_plots} 
\includegraphics[width=0.9\linewidth]{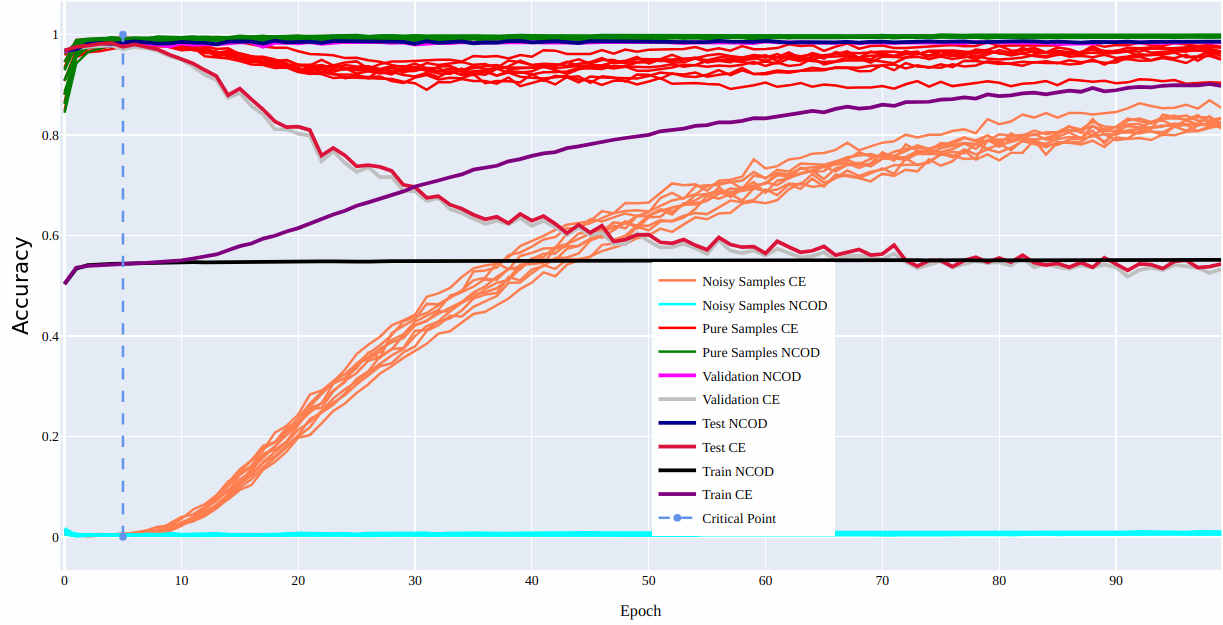}
}\quad
\caption{
The figure shows how our loss function (NCOD) performed compared to Cross entropy (CE).}
\end{figure}

\section{Cross-Dataset Noise Analysis of Loss Functions} \label{appendix:CrossDataset_Noise Analysis_of_Loss_Functions}

To establish the robustness of our analysis across diverse datasets and architectures, we replicated the investigation we did on CIFAR 100 in \cref{fig:train_plots_cifar} on the MNIST dataset using a CNN. 
More precisely the network we used is made of 2 Convolution layers (kernel size equal to 3, number of channels for the first layer is 16 and for the second 32) followed by a MaxPool (kernel size equal to 2) layer and followed by 2 fully connected layers (the first fully connected layer has dimension 128).
Our results reveal a noticeable contrast in training accuracy between pure and noisy samples. Initially, pure samples exhibit superior training accuracy, whereas noisy samples display markedly lower accuracy. Upon employing the cross-entropy approach, we observed a gradual reduction in the disparity of training accuracies as the training proceeded. This phenomenon contributed to a decrease in test accuracy. Conversely, when utilizing our loss function, the gap in training accuracy between pure and noisy samples either remained stable or expanded. This dynamic led to an improvement in test accuracy. For a comprehensive exploration of these outcomes, please consult \cref{fig:train_plots}. The figure presents a comparison of training for CE loss and NCOD loss for the ten different classes in the MNIST dataset. We also plot the test accuracy for both loss functions.

In this figure, there's a specific point labeled as ``critical''. This is where things get interesting. When we use cross-entropy, the training accuracy for nearly all the noisy samples in every class starts to go up at this critical point. This increase in training accuracy, however, leads to a drop in the accuracy of the model's predictions during testing. On the other hand, with our unique loss function (NCOD), it's a different story. This loss function ensures that the training accuracy for noisy samples doesn't increase as it does with the cross-entropy method. Instead, the model primarily learns from clean, noise-free samples as evident from the figure. This special behavior contributes to an overall improvement in the accuracy of the model's predictions during testing.

\end{document}